\documentclass[runningheads]{llncs}

\usepackage[year=2026]{eccv}

\usepackage{eccvabbrv}

\usepackage[abbreviations]{glossaries-extra}
\usepackage{xspace}
\PassOptionsToPackage{table}{xcolor}
\usepackage{colortbl}
\usepackage{array}
\usepackage{xcolor} 
\usepackage{threeparttable}
\usepackage{subcaption}
\usepackage{afterpage}
\usepackage{lipsum}
\usepackage{algorithm}
\usepackage{algpseudocode}  
\usepackage{booktabs}
\usepackage{multirow}
\usepackage{wrapfig}
\usepackage{caption}
\usepackage{amsmath}
\usepackage{sidecap}
\usepackage{xr}
\usepackage{enumitem}
\usepackage{float}
\usepackage{url}

\definecolor{semired}{RGB}{255, 0, 0}
\definecolor{semiorange}{RGB}{255, 165, 0}

\renewcommand{\thefootnote}{\fnsymbol{footnote}}

\newcommand{\refSec}[1]{Sec.~\ref{sec:#1}}
\newcommand{\refSupSec}[1]{Sec.~\ref{supplementary:#1}}
\newcommand{\refSupSecShort}[1]{~\ref{supplementary:#1}}
\newcommand{\refFig}[1]{Fig.~\ref{fig:#1}}
\newcommand{\refFigFull}[1]{Fig~\ref{fig:#1}}

\newcommand{\refTbl}[1]{Table.~\ref{tbl:#1}}

\newabbreviation{HVS}{HVS}{Human Visual System}
\newabbreviation{AR}{AR}{Augmented Reality}
\newabbreviation{VR}{VR}{Virtual Reality}
\newabbreviation{SLM}{SLM}{Spatial Light Modulator}
\newabbreviation{FoV}{FoV}{Field Of View}
\newabbreviation{HOE}{HOE}{Holographic Optical Element}
\newabbreviation{CNN}{CNN}{Convolutional Neural Network}
\newabbreviation{PSF}{PSF}{Point-Spread Function}
\newabbreviation{POH}{POH}{Phase-only Hologram}
\newabbreviation{MLP}{MLP}{Multilayer Perceptron}
\newabbreviation{CBAM}{CBAM}{Convolutional Block Attention Module}
\newabbreviation{FPN}{FPN}{Feature Pyramid Network}
\newabbreviation{MDE}{MDE}{Monocular Depth Estimation}
\newabbreviation{PSP}{PSP}{Pyramid Spatial Pooling}
\newabbreviation{STE}{STE}{Straight-Through Estimator}
\newabbreviation{BCP}{BCP}{Bin Center Predictor}
\newabbreviation{CGH}{CGH}{Computer-Generated Holography}
\newabbreviation{SA}{SA}{Segment Anything}
\newabbreviation{HDR}{HDR}{High Dynamic Range}
\newabbreviation{LR}{LR}{Learning Rate}
\newabbreviation{TV}{TV}{Total Variation}
\newabbreviation{SILog}{SILog}{Scale Invariant Log}
\newabbreviation{MTL}{MTL}{Multi-task Learning}
\newabbreviation{ASM}{ASM}{Angular Spectrum Method}
\newabbreviation{BLASM}{BLASM}{Band-limited Angular Spectrum Method}
\newabbreviation{ViT}{ViT}{Vision Transformer}
\newabbreviation{2D}{2D}{Two-Dimensional}
\newabbreviation{FPS}{FPS}{Frames Per Second}
\newabbreviation{1D}{1D}{One-Dimensional}
\newabbreviation{DP}{DP}{Double Phase}
\newabbreviation{GM}{GM}{Gradient Matching}
\newabbreviation{KD}{KD}{Knowledge Distillation}
\newabbreviation{SAM}{SAM}{Skip Attention Module}
\newabbreviation{SOTA}{SOTA}{state-of-the-art}
\newabbreviation{fp32}{fp32}{32-bit precision}
\newabbreviation{3DGS}{3DGS}{3D Gaussian Splatting}
\newabbreviation{NeRF}{NeRF}{Neural Radiance Fields}
\newabbreviation{threeD}{3D}{Three-Dimensional}
\newabbreviation{SH}{SH}{Spherical Harmonic}
\newabbreviation{FFT}{FFT}{Fast Fourier Transform}
\newabbreviation{SBP}{SBP}{Space-Bandwidth Product}
\newabbreviation{DPAC}{DPAC}{Double Phase-Amplitude Coding}
\newabbreviation{OIT}{OIT}{Order Independent Transparency}
\newabbreviation{GWS}{GWS}{Gaussian Wave Splatting}
\newabbreviation{INR}{INR}{Implicit Neural Representation}
\newabbreviation{OOM}{OOM}{Out of Memory}

\global\long\def\OOM{\gls{OOM}\xspace}
\global\long\def\POH{\gls{POH}\xspace}
\global\long\def\BLASM{\gls{BLASM}\xspace}
\global\long\def\INR{\gls{INR}\xspace}
\global\long\def\GWS{\gls{GWS}\xspace}

\global\long\def\DPAC{\gls{DPAC}\xspace}

\global\long\def\threeD{3D\xspace}

\global\long\def\3DGS{\gls{3DGS}\xspace}

\global\long\def\SLM{\gls{SLM}\xspace}

\global\long\def\CGH{\gls{CGH}\xspace}

\global\long\def\1D{\gls{1D}\xspace}
\global\long\def\2D{\gls{2D}\xspace}
\global\long\def\threeD{\gls{threeD}\xspace}

\global\long\def\fp32{\gls{fp32}\xspace}

\usepackage[accsupp]{axessibility}

\usepackage{hyperref}

\usepackage{orcidlink}
\usepackage{colortbl}
\definecolor{lightorange}{rgb}{1,0.9,0.7}
\definecolor{lightred}{rgb}{1,0.4,0.4}
\definecolor{lightgreen}{rgb}{0.6,1,0.3}
\definecolor{forestgreen}{rgb}{0.133, 0.545, 0.133}
\definecolor{lightyellow}{rgb}{1,1.0,0.6}

\begin{document}

\title{Complex-Valued 2D Gaussian Representation for Computer-Generated Holography}

\titlerunning{Complex-Valued 2D Gaussians for CGH}

\author{Yicheng Zhan\textsuperscript{$\star$}\inst{1}\orcidlink{0009-0006-5936-1929} \and
Xiangjun Gao\textsuperscript{$\star$}\inst{2}\orcidlink{0009-0003-6177-4413} \and
Long Quan\inst{2}\orcidlink{0000-0002-0329-9437} \and
Kaan Akşit\inst{1}\orcidlink{0000-0002-5934-5500}}

\authorrunning{Y.~Zhan et al.}

\institute{University College London, London, UK \and
Hong Kong University of Science and Technology (HKUST), Hong Kong, China\\
\email{yicheng\_zhan2001@outlook.com}}

\maketitle
\let\thefootnote\relax\footnotetext{\textsuperscript{$\star$}~Equal Contribution.}

\begin{abstract}
Complex-valued Gaussian primitives have recently been explored for
representing holographic radiance fields in 3D novel view synthesis.
In this work, we extend this line of research to the hologram optimization domain
and propose a structured representation based on complex-valued 2D Gaussian primitives.
Inspired by Gabor's theory, we show that our primitive attains the minimum space--frequency uncertainty 
and reduces the parameter search space by 5:1 compared to per-pixel parameterization.
To enable end-to-end training, we develop a differentiable rasterizer for our representation,
integrated with a GPU-optimized light propagation kernel in free space.
Extensive experiments show that our method reduces VRAM usage by up to 30\% and accelerates optimization by 50\% over standard 
autodiff-based implementations, delivers up to 13~dB higher PSNR than prior Gaussian-based methods, 
and achieves up to $3200\times$ faster rendering while maintaining 
reconstruction quality on par with existing CGH approaches.
For evaluation, we introduce a conversion procedure that adapts
our representation to practical hologram formats, including smooth and random phase-only holograms.
By reducing the hologram parameter search space, our representation enables a more scalable hologram 
estimation in the next-generation computer-generated holography systems. 
\end{abstract}
\section{Introduction}
\label{sec:intro}

\begin{figure}[!thp]
    \centering
    \includegraphics[width=0.6\columnwidth]{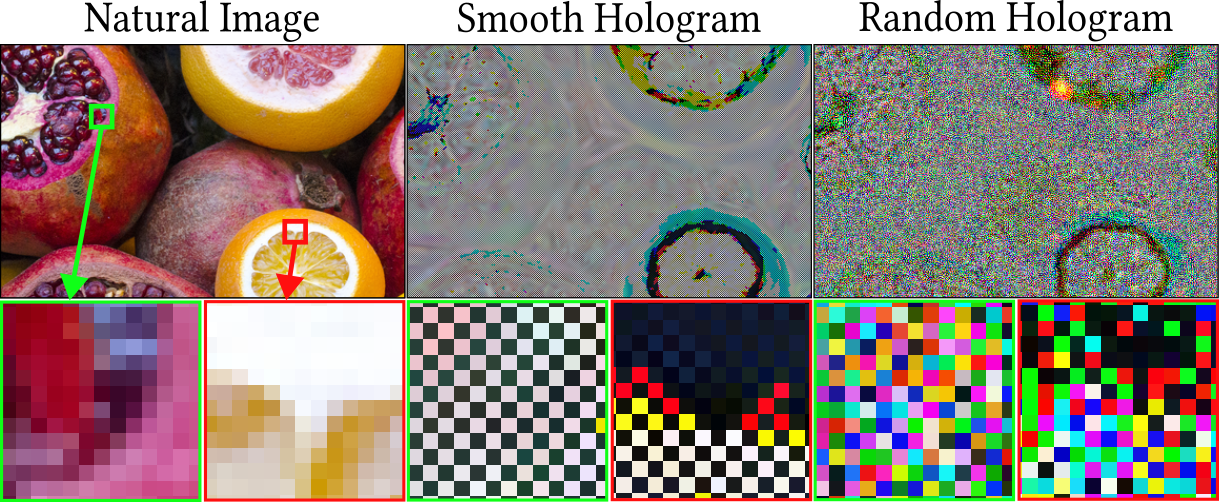}
    \caption{Comparison between a natural image and different hologram formats.
    Unlike the smoother pixels in natural images, holograms produce dense high-frequency and random spatial variations that are challenging to represent
    (Source Image:~\cite{pomegranate}).}
    \label{fig:priliminary}
\end{figure}

Holographic displays are a promising technology for
realistic \threeD content presentations~\cite{kim2024holographic}.
Unlike natural images, which record only light intensity, holograms
capture intensity, interference, and diffraction phenomena.
As shown in~\refFigFull{priliminary}, compared to natural images, holograms exhibit markedly different spatial characteristics.
Therefore, a key challenge in \CGH is to design compact and efficient representations that preserve
high-frequency details of holograms while being scalable~\cite{wang2022joint, Shi2022OL}.
\setlength{\intextsep}{1.5pt}
\setlength{\columnsep}{4pt}

Conventional image representation methods, such as \INR{}~\cite{sitzmann2020implicit, rumelhart1986learning, Peng2025Poster},
optimize a continuous implicit function to represent an image.
However, implicit functions generally favor continuous low-frequency data,
making it difficult to capture the high-frequency details typical of holograms.
In parallel, autoencoder-based approaches~\cite{bohan2023tiny, wang2024sinsr, preechakul2022diffusion, song2023consistency},
typically pretrained on natural images, rely on learned priors that do not readily generalize to hologram structures,
hinting a need for hologram specific adaptations.
Additionally, emerging Gaussian-based image representations~\cite{zhang2024gaussianimage, zhang2025image, Zeng_2025_ICCV}
have recently been proposed for natural image modeling.
While effective for natural images, these methods directly encode hologram pixels
without modeling interference and diffraction phenomena,
leaving room for more specialized hologram representations and better reconstruction quality.

More recently, Gaussian primitives have also been adopted in \CGH for 3D novel view synthesis~\cite{zhan2025complex, chao2025random, choi2025GWS}.
In this work, we extend this line of research to further
discover the potential of Gaussians in the hologram optimization domain.
Instead of modeling a 3D scene from multi-view images,
we focus on representing a \emph{single} hologram through complex-valued 2D Gaussian primitives
for efficient and scalable hologram estimation.
Inspired by Gabor's theory~\cite{gabor1946theory},
we show that our primitive attains the minimum space--frequency uncertainty,
making it the compact and efficient primitive for hologram representation.
To enable end-to-end optimization, as the computational backbone of our method,
we develop a differentiable rasterizer for complex-valued 2D Gaussians,
integrated with a GPU-optimized light propagation kernel, enabling efficient and scalable hologram optimization and rendering.
Our comprehensive evaluation shows that, our method reduces the parameter search space by a \textbf{5:1}
ratio compared to dense per-pixel parameterization (the conventional hologram optimization setting),
decreases VRAM usage by \textbf{30\%}, and accelerates optimization by \textbf{50\%}
compared to standard autodiff-based developments (PyTorch~\cite{NEURIPS2019_9015}),
while achieving reconstruction fidelity comparable to prior Gaussian-based and CGH methods.
Our contribution is summarized as follows:
\begin{itemize}
\item We migrate the complex-valued 3D Gaussians from CGH 3D novel view synthesis~\cite{zhan2025complex}
to \textbf{complex-valued 2D Gaussian} prmitives.
We show the effective usage of this primitive in hologram optimization
by reducing the parameter search space by \textbf{5:1} while preserving reconstruction fidelity
(degrading gracefully up to a \textbf{10:1} upper bound).
\end{itemize}
Additionally, for evaluation purposes, we introduce a conversion procedure that adapts
our complex-valued 2D Gaussian representation to practical hologram formats, including Smooth \POH via double-phase
coding~\cite{DoublePhase} and Random \POH via structural guided phase optimization.
We hope this work demonstrates the broader potential of Gaussian primitives in \CGH and
encourages further investigation into compact, scalable hologram optimization frameworks.
For readers less familiar with the underlying principles of \CGH,
we present the essential concepts in \refSec{Preliminary}. Our code is available at~\cite{ourcode2026}.

\section{Related Work}
\label{sec:related_work}

\subsection{Gaussian and Learned Image Formation}
Recent advances in natural image representation have explored compact alternatives to dense pixel-wise parameterizations.
Autoencoder-based methods~\cite{preechakul2022diffusion, bohan2023tiny, wang2024sinsr} and \INR{}s~\cite{sitzmann2020implicit, saragadam2023wire, lindell2022bacon, rumelhart1986learning}
map natural images into latent spaces or continuous functional representations, enabling structured compression but often favoring smooth, low-frequency content.
More recently, building on \3DGS and its variants~\cite{kerbl20233d, gao2025mani, mallick2024taming, wu20244d},
several works have extended Gaussian primitives from neural rendering to natural image encoding~\cite{zhang2024gaussianimage, Zeng_2025_ICCV, zhang2025image},
leveraging 2D Gaussians for efficient image rendering and compression.
While these Gaussian-based methods show promising results on representing natural images,
they do not model interference and diffraction phenomena,
and are therefore not directly suited for hologram representation.
In this paper, we explore and propose complex-valued 2D Gaussian primitives as a hologram representation
that explicitly integrates light propagation during optimization,
targeting both parameter search space reduction and high-quality hologram reconstruction.

\subsection{Preliminary Concepts: \CGH}
\label{sec:Preliminary}
\setlength{\intextsep}{1.5pt}
\setlength{\columnsep}{5pt}
\begin{wrapfigure}{r}{0.6\columnwidth}
    \centering
    \includegraphics[width=0.58\columnwidth]{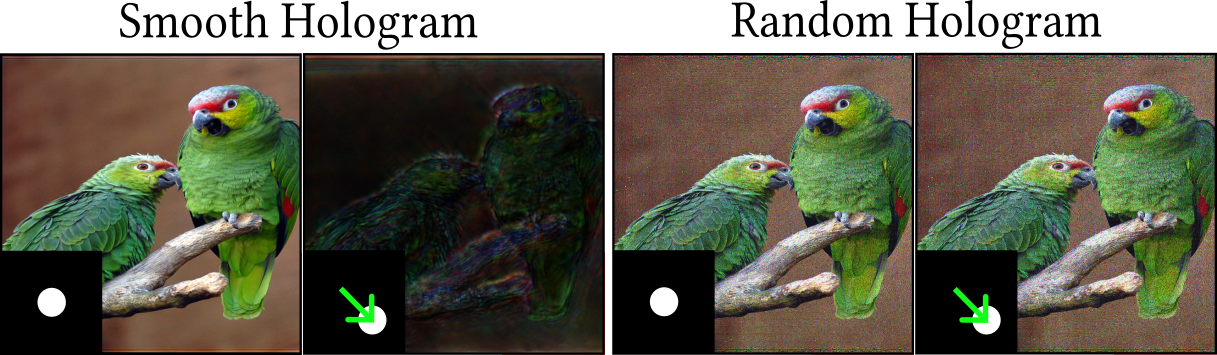}
    \caption{(Simulated) Smooth hologram shows high quality at the pupil center
    but degrade severely with pupil shifts, whereas random hologram remains visible (Source Image:~\cite{Wilson2009}).}
    \label{fig:pupil_offset}
\end{wrapfigure}
\CGH is a \textbf{computational imaging} task that performs wave-based rendering,
synthesizing \textbf{holograms} to reconstruct \threeD scenes for holographic displays,
where the display is typically implemented as \SLM as the programmable wavefront device.
Unlike natural images that capture only smooth, low-frequency intensity variations,
a hologram simultaneously encodes light's intensity, interference, and diffraction.
As shown in~\refFigFull{priliminary}, this wave-optical data format manifests as dense,
high-frequency, and random-valued structures that challenge conventional image representations.
Holographic data can be represented as a complex hologram: $\mathbf{H} = A \exp(j\varphi)$ with amplitude $A$ and phase $\varphi$,
requiring specialized modulators; or as a phase-only hologram: $\mathbf{H}_{\text{POH}} = \exp(j\varphi)$, compatible with commercial holographic displays.
\POH variants include Smooth \POH with spatial multiplexed phase~\cite{DoublePhase} and Random \POH with directly optimized phase~\cite{schiffers2023stochastic}.
For more details of different hologram formats, please refer to \refSec{parallel_training}.

\paragraph{Light Propagation.}
\label{sec:related_propagation}
\setlength{\intextsep}{1.5pt}
\setlength{\columnsep}{5pt}
\begin{wrapfigure}{r}{0.4\columnwidth}
    \centering
    \includegraphics[width=0.38\columnwidth]{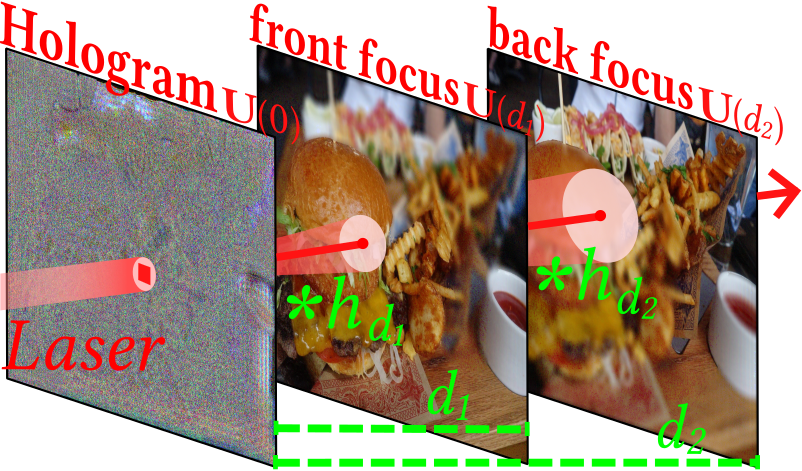}
    \caption{Hologram reconstruction via free-space light propagation (Source Image:~\cite{Burger2014}).}
    \label{fig:holo_propagation}
\end{wrapfigure}
The core of \CGH is free-space propagation based on scalar diffraction theory~\cite{goodman2005Fourier}. As shown in~\refFigFull{holo_propagation},
free-space propagation over a distance $d$ can be expressed as a 2D convolution of the source field $\mathbf{U}(0)$ with the spatial impulse response $h_d$,
\begin{equation}
\mathbf{U}(d) = \mathbf{U}(0) * h_d,
\end{equation}
where $0$ denotes the source plane at zero distance. Equivalently, in the frequency domain
\begin{equation}
\mathbf{U}(d) = \mathcal{F}^{-1}\!\left\{ H_d(f_x,f_y)\,\mathcal{F}\{\mathbf{U}(0)\}\right\},
\end{equation}
where $H_d(f_x,f_y) = \mathcal{F}\{h_d\}$ is the transfer function~\cite{matsushima2009band, Chuanjun2024SigAsia}.
A common choice of $H$ is the \BLASM
\begin{equation}
H_d(f_x,f_y) =
\begin{cases}
\exp\!\left(j\,2\pi d \sqrt{\tfrac{1}{\lambda^2}-r^2}\right), & r^2 \le \tfrac{1}{\lambda^2},\\[6pt]
0, & \text{else},
\end{cases}
\end{equation}
where $r^2 = f_x^2 + f_y^2$ and $\lambda$ denotes the working wavelength.
In this paper, we further optimized \BLASM kernel to model light propagation,
being 50\% faster and 30\% VRAM-efficient than PyTorch development.

\subsection{Holographic Representation}
Conventional \CGH either optimizes per-pixel holograms~\cite{kavakli2023multicolor, kavakli2023realistic, kuo2023multisource, Schiffers2025multiwavelength, Brian2024SigAsia},
or trains neural networks to directly predict hologram pixels~\cite{shi2022end, peng2020neural, shi2021towards, zhan2024Configure, Choi2021Neural3D}, both of which yield large solution spaces and hinder scalability.
More recently, Gaussian primitives have been introduced to bridge computer graphics and holography.
Here we clarify the positioning of our work relative to these efforts.
\emph{3D novel-view-synthesis (NVS) CGH} methods operate on multi-view inputs:
\GWS~\cite{choi2025GWS} and Random-phase \GWS~\cite{chao2025random} leverage pretrained 2DGS scenes~\cite{huang20242d} for geometry-aware modeling of interference and diffraction,
while complex-valued holographic radiance fields~\cite{zhan2025complex} demonstrate that 3D Gaussians can directly represent volumetric holographic scenes.

In contrast, \emph{natural-image Gaussian} representations~\cite{zhang2024gaussianimage, Zeng_2025_ICCV, zhang2025image} contain no wave optics and underperform on holograms,
whereas \emph{per-pixel CGH}~\cite{Choi2021Neural3D, Wirtinger2019, kavakli2023multicolor} is dense and incurs a large solution space.
A concurrent effort further applies 2D Gaussians to compress double-phase holograms~\cite{fan2026compressing}.
Distinct from all of these, our method is a compact complex-valued representation designed specifically for single-view hologram optimization and rendering,
that explicitly integrates light propagation during optimization,
achieving both parameter search space reduction and competitive reconstruction fidelity.

\section{Method}
\label{sec:method}
\setlength{\intextsep}{1.5pt}
\setlength{\columnsep}{5pt}
\begin{figure*}[!htbp]
    \centering
    \includegraphics[width=\textwidth]{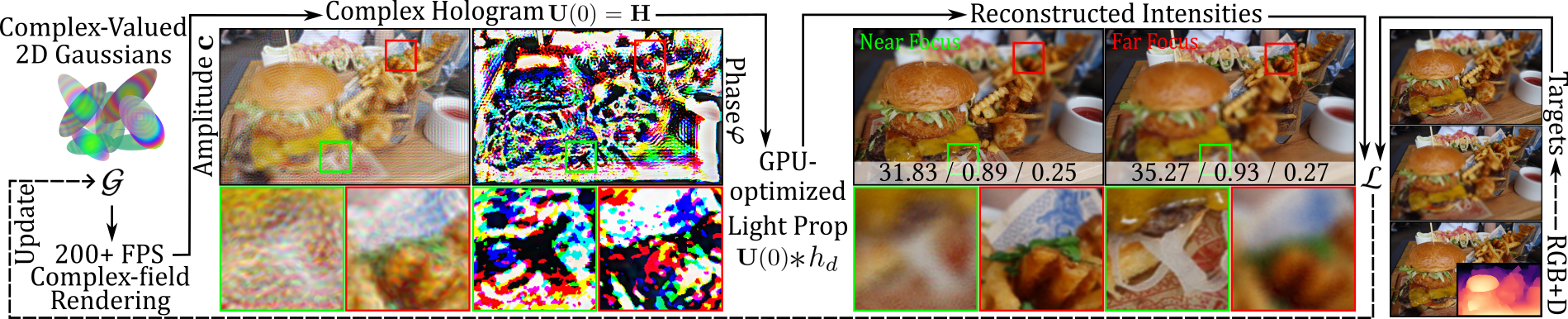}
    \caption{
        Overview of our pipeline.
        Complex-valued 2D Gaussians are rasterized into a complex hologram (amplitude and phase),
        which is propagated to multiple depth planes using optimized light propagation.
        Reconstructions are compared with RGB{+}D derived targets at different focal distances, and we report PSNR, SSIM, and LPIPS.
        Here, \textit{Light Prop} denotes light propagation
        (Source Image: \cite{Burger2014}.)
    }
    \label{fig:system}
\end{figure*}

\paragraph{Problem Definition}
Given a target image $\mathbf{I}_{\text{target}} \in \mathbb{R}^{C \times H \times W}$ and an optional depth map $\mathbf{D} \in \mathbb{R}^{H \times W}$,
we aim to synthesize a complex 3D hologram $\mathbf{H} \in \mathbb{C}^{C \times H \times W}$ whose optical propagation reconstructs $\mathbf{I}_{\text{target}}$ with correct focus and defocus.
For $C=3$ (RGB), each channel of $\mathbf{H}$ is complex-valued (real and imaginary parts), equivalently $\mathbf{H} \in \mathbb{R}^{6 \times H \times W}$.

\subsection{Complex-Valued 2D Gaussian Primitives}
\refFigFull{system} shows the training pipeline of our method.
Building on the development of complex-valued holographic radiance fields~\cite{zhan2025complex},
we extend the representation to define a complex-valued 2D Gaussian primitive for single-hologram optimization.
Whereas~\cite{zhan2025complex} models volumetric 3D scenes from multi-view images,
our formulation operates on the hologram plane directly, targeting efficient and scalable hologram estimation.
Each primitive $\mathcal{G}_n$ is parameterized as
\begin{equation}
\mathcal{G}_n = \{\tilde{\mathbf{x}}_n, \tilde{\mathbf{s}}_n, \theta_n, \mathbf{c}_n, \boldsymbol{\varphi}_n, \tilde{\alpha}_n\},
\end{equation}
where $\tilde{\mathbf{x}}_n \in \mathbb{R}^2$ denotes the pre-activation 2D position, $\tilde{\mathbf{s}}_n \in \mathbb{R}^2$ the pre-activation scales,
$\theta_n \in \mathbb{R}$ the in-plane rotation angle, $\mathbf{c}_n \in \mathbb{R}^C$ the color amplitudes, $\tilde{\alpha}_n \in \mathbb{R}$ the pre-activation opacity,
and $\boldsymbol{\varphi}_n \in \mathbb{R}^C$ the per-channel phase.
Naively, a complex-valued field can be represented by pairing two real-valued Gaussians for the real part and the imaginary part, respectively.
This demands $18$ parameters per primitive pair and two separate renderings,
one for each real-valued Gaussian, resulting in extra computation.
By contrast, our formulation has $12$ parameters, achieving a $\tfrac{1}{3}$ reduction in parameterization while requiring a single rendering.
We keep the amplitude factored as $\alpha_n \mathbf{c}_n$ rather than a single value (ablated in Suppl~\refSupSec{decompose}).
We apply activation functions to enforce valid parameter ranges (See Suppl \refSupSec{activation} for activation formulations).
In the following equations, $\mathbf{x}_n$, $\mathbf{s}_n$, and $\alpha_n$ denote the activated parameters obtained from their corresponding pre-activation counterparts $\tilde{\mathbf{x}}_n$, $\tilde{\mathbf{s}}_n$, and $\tilde{\alpha}_n$.
The spatial distribution of the Gaussian is defined by a 2D covariance matrix $\boldsymbol{\Sigma}_n = \mathbf{R}(\theta_n)\mathbf{S}_n^2\mathbf{R}(\theta_n)^\top$,
where $\mathbf{R}(\theta_n)$ is the rotation matrix and $\mathbf{S}_n = \text{diag}(\mathbf{s}_n)$.
The inverse covariance $\boldsymbol{\Sigma}_n^{-1}$ is computed analytically (see Suppl \refSupSec{2d_cov} for the full covariance calculation).
The contribution of $\mathcal{G}_n$ at the pixel coordinate $\mathbf{p}$ is
\begin{equation}
g_n(\mathbf{p}) = \exp\!\left(-\tfrac{1}{2}(\mathbf{p} - \mathbf{x}_n)^\top \boldsymbol{\Sigma}_n^{-1} (\mathbf{p} - \mathbf{x}_n)\right),
\end{equation}
and the complex-valued hologram pixel at $\mathbf{p}$ is
\begin{equation}
\mathbf{H}_n(\mathbf{p}) = \alpha_n \mathbf{c}_n \, g_n(\mathbf{p}) \exp\!\big(j \boldsymbol{\varphi}_n\big).
\end{equation}
The hologram is formed by the accumulation of all primitives in pixel grid
\begin{equation}
\mathbf{H} = \Bigl\{\sum_{n=1}^{N} \mathbf{H}_n(\mathbf{p}) \;\Big|\; \mathbf{p}\in[1,W]\times[1,H]\Bigr\}.
\end{equation}

\begin{figure}[ht!]
    \centering
    \includegraphics[width=0.9\columnwidth]{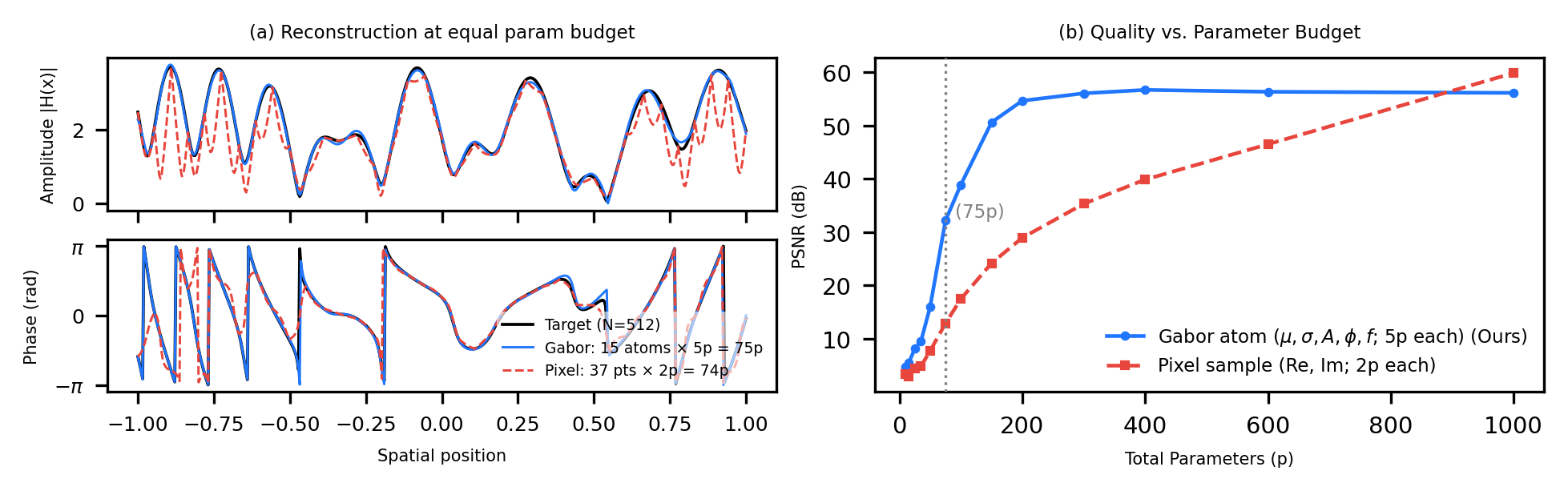}
    \caption{Thought experiment on a complex-valued 1D signal: Gabor atoms versus uniform pixel samples at equal parameter budgets. At matched budgets, Gabor atoms retain higher PSNR, most notably as the budget shrinks.}
    \label{fig:gabor_thought}
\end{figure}

\paragraph{Connection to Gabor's Theory.}
Each pixel on the \SLM diffracts light into propagation angles.
The Fourier transform of the hologram field maps these pixel values to spatial frequencies, defining the reconstructed image.
In holography, we are capturing both spatial (where light originates) and frequency information (the angles at which it propagates) simultaneously.
Gabor's uncertainty principle~\cite{gabor1946theory} states that the more precisely we know one of these properties,
the less precisely we can know the other---a fundamental constraint for all wave-based systems.
For hologram representation, we therefore seek functions that minimize the joint uncertainty between space and frequency domains.
The Gaussian is the unique function achieving this theoretical minimum bound $\Delta x\,\Delta f_x = \tfrac{1}{2}$~\cite{gabor1946theory}.
To illustrate, a rectangular aperture produces a sinc-function Fourier transform with high-frequency oscillations, whereas a Gaussian aperture yields a Gaussian Fourier transform with minimal space--frequency spread.
Our primitive $\mathbf{H}_n(\mathbf{p})=\alpha_n\mathbf{c}_n\,g_n(\mathbf{p}) \exp(j\boldsymbol{\varphi}_n)$
is the two-dimensional analogue of Gabor's 1D \textbf{elementary signal}~\cite{gabor1946theory},
$\psi(x)=\exp(-\beta^{2}(x-x_{0})^{2})\cdot\exp(j(2\pi f_{0}x+\boldsymbol{\varphi}))$.
Since a 2D Gaussian factorizes along its principal axes,
it attains minimum space--frequency uncertainty in both dimensions,
meaning fewer primitives are needed to faithfully represent a given hologram.
Note that high-frequency content in our reconstructed field does not arise from variations within individual primitives,
but from the superposition of $N$ primitives at positions $\mathbf{x}_n$
and from the frequency-dependent phase applied by \BLASM{}.
To isolate the benefit of Gabor primitives from the full hologram pipeline,
we conduct a controlled study on a complex-valued 1D signal.
At equal parameter budgets, we reconstruct the signal with Gabor atoms versus uniform pixel samples.
As shown in \refFig{gabor_thought}, Gabor atoms consistently achieve higher PSNR than pixel samples across budgets.
This supports our central claim: Gabor's theory does not assert that Gaussians \emph{surpass} per-pixel quality at full budget,
but that they \emph{retain} quality most effectively as the budget shrinks, owing to their minimum space--frequency uncertainty.
%

\subsection{Hologram Reconstruction and Optimization}
Naively, we can supervise $\mathbf{H}$ with $\mathbf{H}_{target}$, however, this approach is insufficient in practice.
To faithfully represent a hologram, it is essential to explicitly incorporate light propagation during optimization.
Given a hologram field $\mathbf{U}(0)=\mathbf{H}$, we simulate free-space propagation using the convolutional methods summarized in Sec.~\ref{sec:related_propagation}.
Specifically, we adopt the \BLASM~\cite{matsushima2009band} as $H_d$ and compute
$\mathbf{U}(d)=\mathcal{F}^{-1}\!\left\{ H_d^{\mathrm{BLASM}}(f_x,f_y)\,\mathcal{F}\{\mathbf{U}(0)\}\right\}$.
To capture depth-dependent effects, we reconstruct $\mathbf{U}(0)$ on $L$ uniformly spaced parallel planes ${\Pi_l}_{l=1}^L$
(spacing distance $\Delta z$, e.g., 2 mm) along the optical axis, centered at propagation distance $d_0$ (e.g., 5 mm),
yielding $d_1 = d_0 - \tfrac{L-1}{2}\Delta z$, $d_L = d_0 + \tfrac{L-1}{2}\Delta z$,
and the reconstructed intensity at plane $l$ is $I_l \;=\; \big|\mathbf{U}(d_l)\big|^2$.
Naively, we can directly supervise $I$ against the target image $\hat{I}$ per depth plane $l$ using $\mathcal{L}_{MSE} = \frac{1}{L}\sum_{l=1}^{L} \|I_l - \hat{I}_l\|^2$,
where $\hat{I}_l$ denotes the ground-truth multi-plane focal stack images generated by Kavaklı et al.~\cite{kavakli2023multicolor}.
To further improve the defocus region`s image quality,
we utilize the reconstruction loss $\mathcal{L}_{recon}$ by Kavaklı et al.~\cite{kavakli2023realistic}, computed as
\begin{equation}
   \begin{split}
   \mathcal{L}_{recon} &= \frac{1}{L} \sum_{l=1}^{L} ( \|I_l - \hat{I}_l\|^2
   + \|I_l \cdot M_l - \hat{I}_l \cdot M_l\|^2 + \|I_l \cdot \hat{I}_l - \hat{I}_l \cdot \hat{I}_l\|^2 ),
   \end{split}
\end{equation}
where $M_l$ is the binary mask for depth plane $\Pi_l$ generated from the target image and its quantized depth.
We also employ the SSIM loss, the final training loss is $\mathcal{L} = \mathcal{L}_{recon} + \lambda_1 \cdot \mathcal{L}_{SSIM}$, where $\lambda_1 = 0.005$.

\subsection{\POH Conversion Procedure}
\label{sec:parallel_training}
Our complex-valued 2D Gaussians provide a compact, efficient hologram representation in the complex
domain. Although there are prototypes utilizing full complex holograms \cite{10.1145/3130800.3130832},
the commercial holographic displays are predominantly \textbf{Liquid Crystal based} phase-only displays
that do not support displaying complex holograms directly \cite{LAM2021050011}.
To bridge this gap, we design a simple yet effective conversion procedure that adapts our complex representation as
\emph{structural guidance} to different hologram formats, including Smooth \POH and Random \POH.
Rather than optimizing each device-specific format end-to-end, we keep a single complex representation as a general
intermediate convertible to \emph{both} formats, decoupling the representation from hardware-specific encoding.

\paragraph{Smooth \POH.}
Our hologram is represented as
$\mathbf{H} = \alpha \, \mathbf{c} \, g \cdot \exp(j\varphi)$.
We employ \DPAC~\cite{DoublePhase} to convert $\mathbf{H}$ into a smooth, phase-only representation by
spatially multiplexing amplitude $A = \alpha \, \mathbf{c} \, g$ and phase $\boldsymbol{\varphi}$ via a checkerboard pattern:
$\boldsymbol{\varphi}_{\text{DPAC}}(i,j) = A(i,j)$ if $(i+j)$ is even, and $\boldsymbol{\varphi}(i,j)$ if odd.
The converted Smooth \POH is $\mathbf{H}_{\text{smooth}} = \exp(j\boldsymbol{\varphi}_{\text{DPAC}})$.

\paragraph{Random \POH.}
\label{sec:parallel_training_random}
We leverage our complex representation as structural guidance for the Random \POH conversion.
The Random \POH is parameterized as $\mathbf{H}_{\text{rand}} = \exp(j\varphi_{\text{rand}})$,
where $\varphi_{\text{rand}} \in \mathbb{R}^{C \times H \times W}$ denotes learnable randomly-initialized phase values.
Both $\mathbf{H}$ and $\mathbf{H}_{\text{rand}}$ are propagated through the \BLASM in parallel,
obtaining reconstructions at the same depth plane $l$ with
$I_l = |\mathbf{U}(d_l)|^2$ and
$I_{\text{rand}}^{(l)} = |\mathbf{U}_{\text{rand}}(d_l)|^2$.
We jointly optimize the following objectives in both intensity and complex field domains
\begin{equation}
\begin{split}
\mathcal{L}_{\mathrm{extract}} = \sum_{l=1}^{L} \Big[
&\mathcal{L}_{\mathrm{recon}}(I_l, \hat{I}_l)
+ \mathcal{L}_{\mathrm{recon}}(I_{\mathrm{rand}}^{(l)}, \hat{I}_l) \\
&+ \lambda_{\mathrm{comp}} \|I_l - I_{\mathrm{rand}}^{(l)}\|^2
+ \lambda_{\mathrm{field}} \,\|\mathbf{U}(d_l) - \mathbf{U}_{\mathrm{rand}}(d_l)\|_{1,\mathbb{C}} \Big],
\end{split}
\end{equation}
where $\lambda_{\mathrm{comp}} = 0.1$ and $\lambda_{\mathrm{field}} = 0.01$.
Here, $|\cdot|_{1,\mathbb{C}}$ is the sum of the L1-norms of the real and imaginary components.

\subsection{Efficient CUDA Rendering and Propagation}
\label{sec:CUDA}
We develop our hologram representation pipeline using CUDA, covering both complex-valued rasterization and light propagation in the Fourier domain.

\paragraph{Complex-Valued 2D Gaussian Rasterizer.}
We adapt the tile-based rasterizer from complex-valued holographic radiance field~\cite{zhan2025complex} to 2D Gaussians for a \emph{single hologram}.
Each primitive contributes a complex value instead of a real scalar.
The forward pass decomposes Gaussians into real and imag components via trigonometric evaluation.
We retain the $16 \times 16$ tile structure with duplicate-with-keys assignment and radix sorting, but store only per-pixel opacity,
recovering intermediate values during the backward pass by division for constant VRAM overhead.
Gradients of amplitude and phase require trigonometric chain rules with negated sine/cosine terms from the complex exponential derivative.
Position, covariance, and opacity gradients follow \cite{kerbl20233d} and adapted to 2D screen-space.
For the details of development and derivation, please refer to Suppl \refSupSec{2d_cuda_gradient}.

\paragraph{GPU-optimized Light Propagation.}
Additionally, as part of the rasterizer, we develop a \BLASM kernel that processes spatial frequencies in parallel and evaluates the transfer function to simulate light propagation effectively.
Valid frequencies are multiplied by the transfer function using accelerated trigonometric operations.
The backward pass applies the conjugate transfer function while preserving bandlimiting, and achieves efficiency through coalesced read-only cache access.
For the details of development and derivation, please refer to Suppl \refSupSec{light_propagation}.

\section{Implementation}
We initialize $N$ Gaussians by uniformly sampling image-plane positions $\mathbf{x}_n^{\text{raw}} \sim \text{Uniform}([0, W] \times [0, H])$ and transforming 
to the unconstrained domain via $\tilde{\mathbf{x}}_n^{\text{init}} = \operatorname{atanh}(2\mathbf{x}_n^{\text{raw}}/[W, H] - 1)$. 
Scales are set to $\tilde{\mathbf{s}}_n = \log([1.5, 5.0])$ pixels, colors $\mathbf{c}_n$ sampled from $[0,1]$, phases $\boldsymbol{\varphi}_n$ initialized to zero, 
and opacity pre-activations fixed at $\tilde{\alpha}_n = -0.5$ (yielding $\alpha_n \approx 0.38$ post-sigmoid). 
We optimize using Adan~\cite{xie2024adan} with empirically-selected, parameter-specific learning rates: positions $10^{-2}$ (cosine-annealed to $10^{-3}$~\cite{CosineAnnealingLR}), 
scales $5 \times 10^{-3}$, amplitudes and phases $2.5 \times 10^{-3}$, opacities $2.5 \times 10^{-2}$, and rotations $10^{-3}$. 
Training runs for 2000 steps on a single NVIDIA RTX 3090 GPU with a 5:1 parameter reduction ratio, with convergence typically occurring around 1000+ steps 
(training visualizations in Suppl~\refSupSec{training_vis_steps}). 
Depth maps are provided by Depth Anything v2~\cite{depth_anything_v2} and MiDaS~\cite{Ranftl2022}. We adopt wavelengths of $639 \text{nm}$, 
$532 \text{nm}$, and $473 \text{nm}$ for red, green, and blue channels respectively, with pixel pitch $3.74 \mu\text{m}$, propagation distance $3 \text{mm}$, 
and volume depth $4 \text{mm}$, consistent with prior work~\cite{shi2021towards, shi2022end, aksit2023holobeam, zhan2024Configure}. 
Our experiments use a LASOS MCS4 RGB laser, Jasper JD7714 phase-only \SLM ($2400 \times 4094$, $3.74~\mu$m), 
and a lens-based optical relay with spatial filtering.
Reconstructions are recorded by a Point Grey GS3-U3-23S6M-C lensless sensor mounted on a motorized stage.
Hardware details are provided in Suppl~\refSupSec{hardware}.
\section{Evaluation}
\begin{figure}[thp]
    \centering
    \includegraphics[width=0.6\columnwidth]{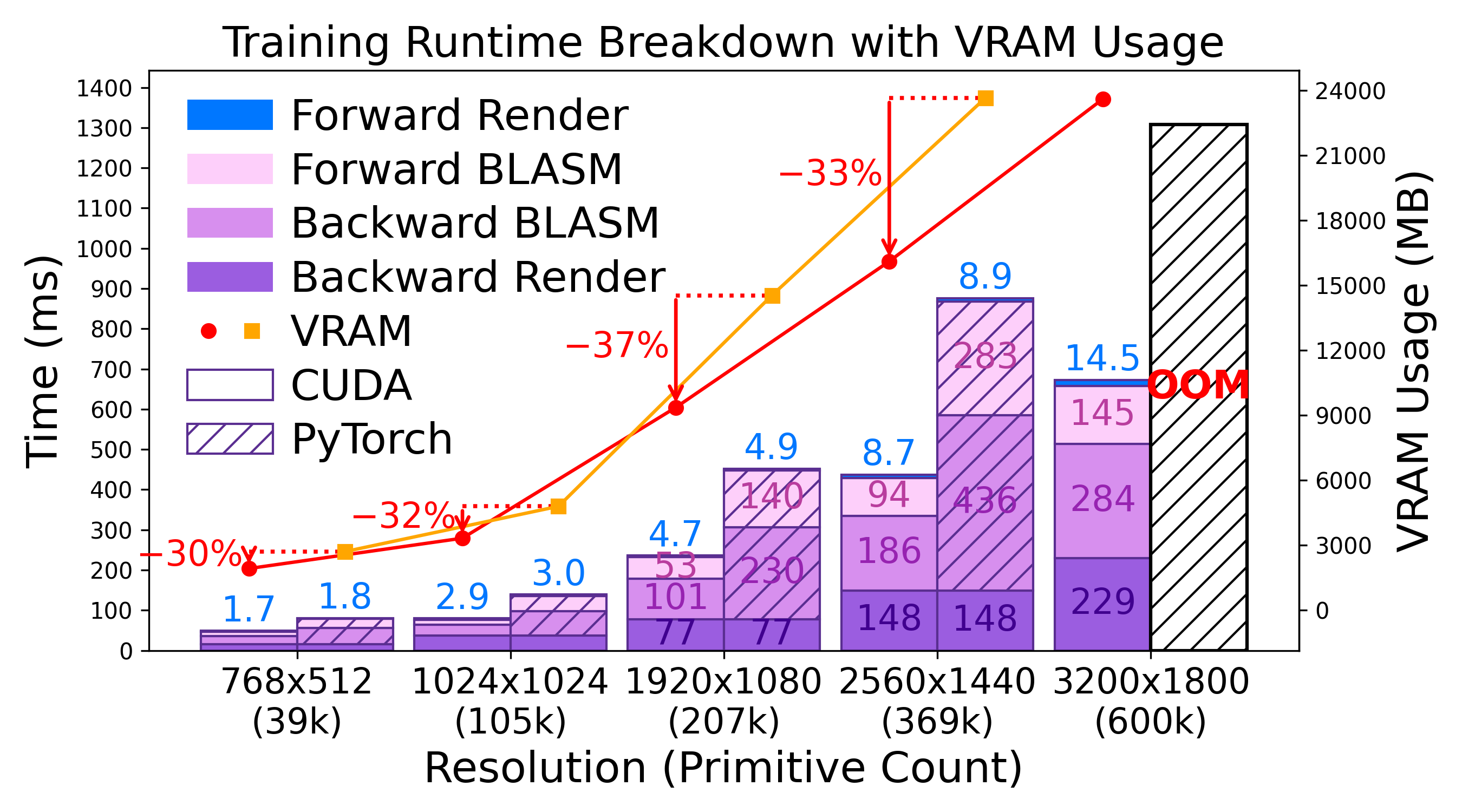}
    \caption{Runtime (bar) and VRAM usage (line) across spatial resolutions for our method ($L = 3$),
    comparing CUDA-based development with the PyTorch baseline. Red downward arrows and percentages indicate the VRAM reduction rate.}
    \label{fig:training_runtime}
\end{figure}
%
%
%
\noindent In this section, we conduct a comprehensive evaluation of our method against baselines.
We use the first $50$ images from DIV2K~\cite{Timofte_2017_CVPR_Workshops} dataset as the test set;
for each method, training and evaluation are performed at a resolution of $3 \times 1024 \times 640$ ($L = 2$).
We report mean PSNR, SSIM, and LPIPS~\cite{zhang2019LPIPS},
averaged over across planes, together with parameter counts, peak VRAM usage, and training time.
For results of training progression, varying depth-plane,
and varying propagation distances, please refer to Suppl~\refSupSec{training_vis_steps},~\refSupSecShort{depth_planes}, and~\refSupSecShort{different_Z}.

\subsection{Runtime and Memory Performance}
\refFigFull{training_runtime} presents the performance results of our kernel; the development detail is provided in ~\refSec{CUDA}.
Across all resolutions, our CUDA kernel consistently reduces VRAM usage by $29$--$36\%$ and accelerates runtime by $40$--$50\%$ compared to the PyTorch developments.
Specifically, memory savings are $28.9\%$, $30.7\%$, $35.6\%$, and $31.9\%$ at $768\times512$, $1024\times1024$, $1920\times1080$, and $2560\times1440$, respectively.
The runtime gains come primarily from the BLASM components: the Forward pass is $47$–$67\%$ faster and the Backward pass is $56$–$58\%$ faster,
while the rasterization stages remain unchanged. Consequently, the overall step time is reduced by $39.7\%$, $42.6\%$, $47.7\%$,
and $50.1\%$, with corresponding memory reductions of $28.9\%$, $30.7\%$, $35.6\%$, and $31.9\%$ at the four resolutions.
With $L = 3$, our method scales up to a resolution of $3200\times1800$ (5.8M pixels) without \OOM,
validating the scalability and efficiency of our method.

\begin{table}[thp]
    \scriptsize
    \setlength{\tabcolsep}{3pt}
    \centering
    \caption{Quantitative comparison of our method, Gaussian-based and learned representation methods.
    Top 2 metrics are highlighted in orange and yellow (* pretrained).}
    \label{tbl:method_GI_TAESD_SIRENcomparison}
\begin{tabular}{lcccccc}
        \toprule
        Method & PSNR $\uparrow$ & SSIM $\uparrow$ & LPIPS $\downarrow$ & VRAM & Params & Time (min) \\
        \midrule
        TAESD*~\cite{bohan2023tiny} & 11.6 & 0.09 & 0.79 & 2.7 G & 2.5 M & - \\
        MLP~\cite{rumelhart1986learning} & 7.5 & 0.04 & 0.85 & 9.9 G & \cellcolor{lightyellow}1.0 M & 6.9 \\
        SIREN~\cite{sitzmann2020implicit} & 7.6 & 0.05 & 0.84 & 13.1 G & \cellcolor{lightyellow}1.0 M & 7.8 \\
        Image-GS~\cite{Zeng_2025_ICCV} & 17.2 & 0.29 & 0.70 & \cellcolor{lightyellow}1.3 G & 2.4 M & 1.6 \\
        Instant-GI~\cite{zhang2025image} & \cellcolor{lightyellow}23.5 & \cellcolor{lightyellow}0.56 & \cellcolor{lightyellow}0.56 & 3.4 G & 2.8 M & \cellcolor{lightyellow}0.9 \\
        GI~\cite{zhang2024gaussianimage} & 22.6 & 0.49 & 0.59 & \cellcolor{lightorange}1.1 G & 2.4 M & \cellcolor{lightorange}0.8 \\
        Ours & \cellcolor{lightorange}\textbf{30.7} & \cellcolor{lightorange}\textbf{0.86} & \cellcolor{lightorange}\textbf{0.33} & 2.2 G & \cellcolor{lightorange}0.8 M & 1.4 \\
        \bottomrule
    \end{tabular}
\end{table}
\begin{figure*}[thp]
    \centering
    \includegraphics[width=1\textwidth]{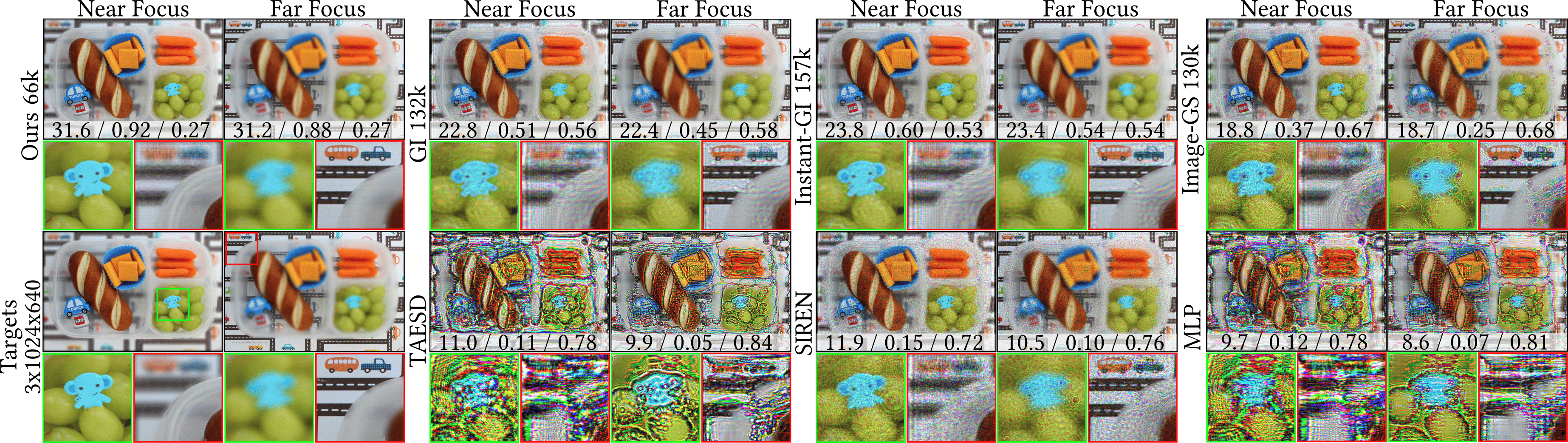}
    \caption{Qualitative comparison of simulated reconstructions at near and far focal planes. Our method uses a 5:1 parameter ratio,
    while existing Gaussian-based approaches~\cite{zhang2024gaussianimage, zhang2025image, Zeng_2025_ICCV} use equal primitive counts for the two real components of the complex field.
    \emph{GI} denotes GaussianImage~\cite{zhang2024gaussianimage}; \emph{Instant-GI} uses network-predicted initialization with variable primitive counts (Source Image:~\cite{AnotherLunch2011, bluecat2012}).}
    \label{fig:TAESD_GI_SIREN_compare}
\end{figure*}

\subsection{Comparison With Representation Methods}

\noindent\textbf{Reconstruction Fidelity.}
\refTbl{method_GI_TAESD_SIRENcomparison} compares our method with Gaussian-based and learned representations for complex hologram modeling.
Learned representations show limited effectiveness: MLP~\cite{rumelhart1986learning} and SIREN~\cite{sitzmann2020implicit} yield
PSNR below $8$ dB and LPIPS around $0.85$, while pretrained TAESD~\cite{bohan2023tiny} reaches $11.6$ dB,
suggesting that \INR{} and autoencoder-based methods are not well-suited for encoding hologram structures.
Gaussian-based methods perform substantially better, with Image-GS~\cite{Zeng_2025_ICCV} achieving $17.2$ dB and
GI~\cite{zhang2024gaussianimage} and Instant-GI~\cite{zhang2025image} reaching $22.6$ and $23.5$ dB,
reflecting the advantage of explicit representations for high-frequency hologram content.
Our method reports $30.7$ dB PSNR, $0.86$ SSIM, and $0.33$ LPIPS.
As shown in \refFigFull{TAESD_GI_SIREN_compare}, prior Gaussian-based methods exhibit structural distortions and blurred details,
while learned approaches introduce severe artifacts and structural loss; in contrast, our method preserves sharper
reconstructions across near- and far-focus planes.

\noindent\textbf{Efficiency and Memory Usage.}
As shown in \refTbl{method_GI_TAESD_SIRENcomparison}, our method uses $0.8$M parameters,
fewer than other Gaussian-based approaches ($2.4$--$2.8$M).
This is due to our compact complex-valued 2D Gaussian definition that avoids the paired real/imaginary parameterization.
In terms of memory, our method requires $2.2$G VRAM, which is moderate due to the explicit incorporation of light propagation;
Gaussian-based methods that omit propagation use less memory but at the cost of fidelity.
Training time remains comparable to Gaussian-based methods
and is considerably shorter than learned approaches.
These results suggest that our representation offers a reasonable trade-off
between reconstruction quality, parameter efficiency, and computational cost for hologram modeling.

\subsection{Comparison With CGH Methods}
\begin{figure*}[!t]
    \centering
    \includegraphics[width=1\textwidth]{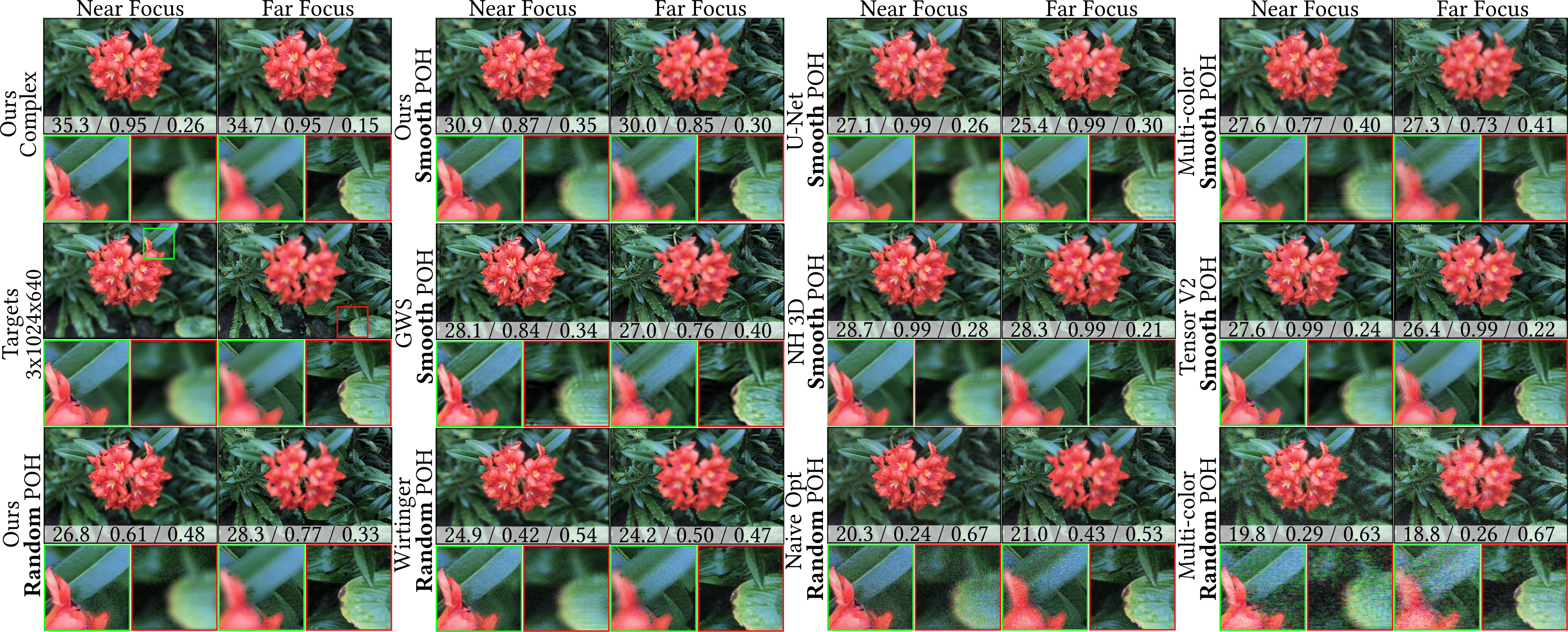}
    \caption{Comparison of simulated reconstructions at near and far focus using different optimization and learned \CGH methods.
    \emph{Multi-color} refers to ~\cite{kavakli2023multicolor};
    \emph{U-Net}~\cite{ronneberger2015u} refers to typical learned \CGH networks widely used in~\cite{liu20234k, hossein2020deepcgh, peng2020neural, Choi2021Neural3D, zhan2024Configure, aksit2023holobeam, chen2025view} (Source Image:~\cite{mildenhall2019llff}).}
    \label{fig:flower_main_performance_compare}
\end{figure*}

\begin{table}[thp]
    \scriptsize
    \setlength{\tabcolsep}{3pt}
    \centering
    \caption{Quantitative comparison across our method, learned CGH and optimization methods.
    (* Naive Opt refers to Random \POH baseline used in~\cite{kuo2023multisource, Schiffers2025multiwavelength, Praneeth2020hitl, peng2020neural}).
    Our Smooth \POH time covers complex optimization only (\POH conversion adds $<$1\,s); Random \POH time includes the conversion.}
    \label{tbl:method_opt_comparison}
    \begin{tabular}{lccccccc}
        \toprule
        Method & PSNR $\uparrow$ & SSIM $\uparrow$ & LPIPS $\downarrow$ & VRAM & Params & Time (min) & Render (ms) \\
        \midrule
        \textbf{Smooth \POH} \\
        NH3D~\cite{Choi2021Neural3D} & \cellcolor{lightyellow}28.3 & \cellcolor{lightyellow}0.92 & \cellcolor{lightyellow}0.31 & 7.1 G & 3.9 M & 90 & 31 \\
        TensorV2~\cite{shi2022end} & 27.1 & \cellcolor{lightorange}\textbf{0.94} & \cellcolor{lightorange}\textbf{0.29} & 8.7 G & \cellcolor{lightorange}0.1 M & 80 & 48 \\
        GWS~\cite{choi2025GWS} & 28.2 & 0.76 & 0.41 & \cellcolor{lightyellow}2.5 G & 1.1 M & \cellcolor{lightyellow}5.1 & 6840 \\
        U-Net~\cite{zhan2024Configure} & 27.2 & 0.91 & 0.35 & 6.3 G & 2.2 M & 100 & \cellcolor{lightyellow}18 \\
        Multi-color~\cite{kavakli2023multicolor} & 27.9 & 0.74 & 0.40 & 3.2 G & 4.0 M & 5.3 & - \\
        Ours & \cellcolor{lightorange}\textbf{29.0} & 0.81 & 0.38 & \cellcolor{lightorange}2.4 G & \cellcolor{lightyellow}0.8 M & \cellcolor{lightorange}1.4 & \cellcolor{lightorange}2.13 \\
        \midrule
        \textbf{Random \POH} \\
        Naive Opt* & 19.8 & 0.33 & 0.60 & \cellcolor{lightorange}2.9 G & \cellcolor{lightorange}2.0 M & \cellcolor{lightyellow}2.9 & - \\
        Wirtinger~\cite{Wirtinger2019} & \cellcolor{lightyellow}25.3 & \cellcolor{lightyellow}0.47 & \cellcolor{lightyellow}0.48 & 3.5 G & \cellcolor{lightorange}2.0 M & \cellcolor{lightorange}2.8 & - \\
        Multi-color~\cite{kavakli2023multicolor} & 20.3 & 0.35 & 0.64 & \cellcolor{lightyellow}3.1 G & 4.0 M & 3.0 & - \\
        Ours & \cellcolor{lightorange}\textbf{29.4} & \cellcolor{lightorange}\textbf{0.81} & \cellcolor{lightorange}\textbf{0.34} & 3.4 G & \cellcolor{lightorange}2.0 M & 3.8 & - \\
        \bottomrule
    \end{tabular}
\end{table}
\refTbl{method_opt_comparison} shows the quantitative results of our method alongside existing \CGH methods for both Smooth and Random \POH.
Additionally, in \refFigFull{flower_main_performance_compare}, we show the qualitative results in an overview and zoomed-in details.
Both Random and Smooth \POH extracted from our
complex-valued 2D Gaussians achieve sharp focus edges and high perceptual quality.

\noindent\textbf{Smooth \POH.} Our method achieves comparable reconstruction quality to learned \CGH methods
(NH3D~\cite{Choi2021Neural3D}, Tensor V2~\cite{shi2022end}, and U-Net~\cite{zhan2024Configure}),
with PSNR of $29.0$ dB, while requiring fewer parameters ($0.8$ M), lower VRAM ($2.4$ G), and shorter training time ($1.4$ min).
Since GWS is a novel view synthesis method, we extend its development to support single-plane \POH rendering.
Compared with GWS~\cite{choi2025GWS}, our method achieves higher fidelity and is substantially faster,
reducing optimization time from $5.1$ to $1.4$ minutes ($3.6\times$)
and render latency from $6840$ to $2.13$ ms ($\sim$3200$\times$ faster).

\noindent\textbf{Random \POH.} Existing baselines operate directly on per-pixel phase and yield limited fidelity,
with Naive Opt and Multi-color~\cite{kavakli2023multicolor} remaining below $21$ dB.
Wirtinger Holography~\cite{Wirtinger2019} reaches $25.3$ dB PSNR; our method reports $29.4$ dB PSNR, $0.81$ SSIM, and $0.34$ LPIPS,
corresponding to gains of $+4.1$ dB PSNR, $+0.34$ SSIM, and $-0.14$ LPIPS over Wirtinger
at comparable memory and modest additional training time ($3.8$ vs.\ $2.8$ min).
These results indicate that structural guidance from complex-valued 2D Gaussians and the \POH conversion procedure
is beneficial for random-valued, per-pixel hologram optimization.

\begin{table}[thp]
    \scriptsize
    \setlength{\tabcolsep}{4pt}
    \centering
    \caption{Comparison of our method under different parameter reduction ratios.}
    \label{tbl:parameter_ratio}
    \begin{tabular}{lccccc}
        \toprule
        Parameter Reduction Ratio & Params & PSNR $\uparrow$ & SSIM $\uparrow$ & LPIPS $\downarrow$ & Render (ms)\\
        \midrule
        Dense Per-pixel & 4.0 M & \cellcolor{lightorange}\textbf{32.3} & \cellcolor{lightorange}\textbf{0.893} & \cellcolor{lightorange}\textbf{0.29} & 4.02 \\
        2:1 & 2.0 M & \cellcolor{lightyellow}31.9 & \cellcolor{lightyellow}0.891 & \cellcolor{lightyellow}0.30 & 2.58 \\
        3:1 & 1.3 M & 31.5 & 0.885 & 0.31 & 2.33 \\
        5:1 & 0.8 M & 30.7 & 0.863 & 0.33 & 2.13 \\
        7:1 & 0.6 M & 30.3 & 0.856 & 0.34 & \cellcolor{lightyellow}1.90 \\
        10:1 & 0.4 M & 29.4 & 0.835 & 0.37 & \cellcolor{lightorange}1.72 \\
        \bottomrule
    \end{tabular}
\end{table}
\subsection{Ablation Study}
\begin{figure*}[!t]
    \centering
    \includegraphics[width=1\textwidth]{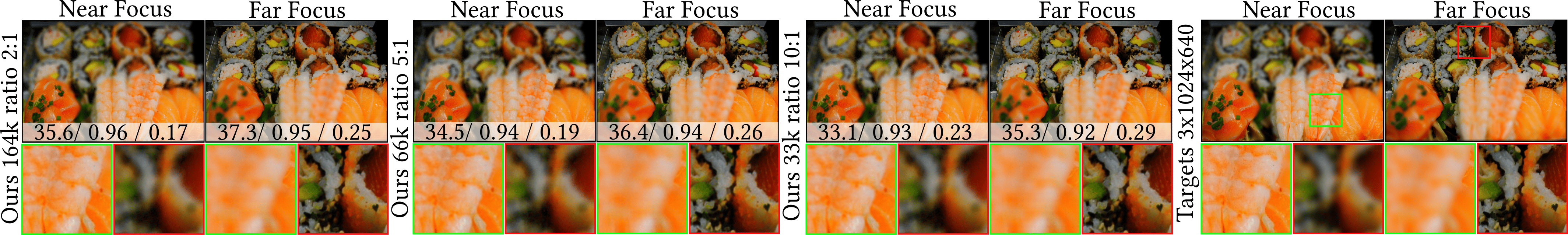}
    \caption{Simulated reconstructions with varying Gaussian counts.
    Labels (e.g., \emph{Ours 33k ratio 10:1}) denote the number of complex 2D Gaussians (33k)
    and the parameter reduction ratio compared to dense per-pixel representation (10:1)
    (Source Image:~\cite{Sutherland2010}).}
    \label{fig:different_param_ratio}
\end{figure*}

\noindent\textbf{Parameter Space Reduction.}
\refTbl{parameter_ratio} evaluates our method under different parameter reduction ratios.
We adopt $5{:}1$ as the default used in all main tables, while $10{:}1$ serves as an ablation upper bound rather than a headline claim.
Quality degrades gracefully: PSNR remains highly comparable at $2{:}1$ ($31.9$) and declines only moderately to $30.7$ and $29.4$ at $5{:}1$ and $10{:}1$, while rendering accelerates accordingly.
\refFigFull{different_param_ratio} confirms that reconstructions remain sharp under aggressive reduction.
Consistent with Gabor's theory, our primitive does not surpass per-pixel quality but \emph{retains} it most effectively as parameters decrease,
making it an efficient basis for compact hologram representation.
\begin{table}[thp]
    \scriptsize
    \centering
    \setlength{\tabcolsep}{3pt}
    \caption{Ablation study of POH conversion losses and Gaussian definition choices, evaluated on the test image~\cite{dragon2016} and~\cite{spaceship2020}.}
    \label{tbl:ablation_poh}
    \begin{tabular}{llccccc}
        \toprule
        Type & Method & PSNR $\uparrow$ & SSIM $\uparrow$ & LPIPS $\downarrow$ & Params & Render (ms)\\
        \midrule
        \multirow{2}{*}{\textbf{Random \POH}} & w/o guidance & 19.1 & 0.37 & 0.52 & 2.0 & -\\
        & with guidance (Ours) & 30.6 & 0.88 & 0.22 & 2.0 & -\\
        \midrule
        \multirow{2}{*}{\textbf{Complex}} & Naive Paired Gaussians & 25.5 & 0.74 & 0.47 & 1.2 & 20.1 \\
        & Ours & 31.8 & 0.89 & 0.31 & 0.8 & 2.13 \\
        \bottomrule
    \end{tabular}
\end{table}
\begin{figure}[thp]
    \centering
    \includegraphics[width=0.99\columnwidth]{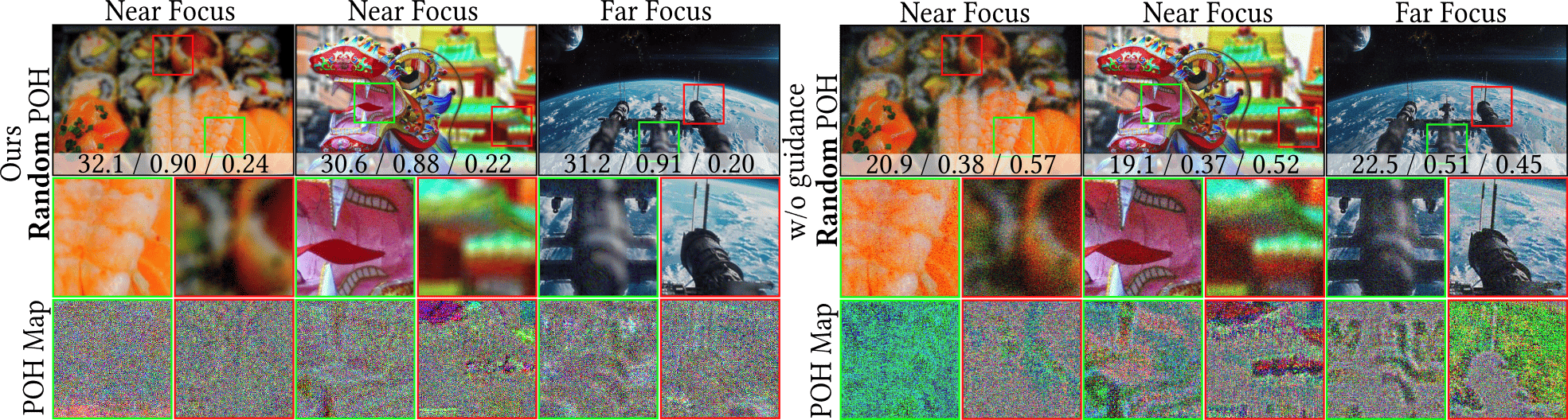}
    \caption{Simulated reconstructions of our method and independently optimized Random \POH at near and far focal planes,
    using identical training strategies. Insets show corresponding hologram pixels
    (Source Image: ~\cite{Sutherland2010, dragon2016, spaceship2020}).}
    \label{fig:distill}
\end{figure}
\noindent\textbf{Random \POH Conversion Quality.}
\refTbl{ablation_poh} compares our \POH conversion to independently optimized Random \POH.
With structural guidance and a more constrained search space, it improves reconstruction quality by $+11.5$ dB.

\textit{Simulation.} \refFigFull{distill} reports quantitative results in simulation.
Our conversion procedure yields higher-quality reconstructions and, as a side effect,
appears to reduce the noise artifacts commonly seen in Random \POH methods~\cite{kavakli2023multicolor, Chu2025RealTime, kuo2023multisource}.
The zoomed insets in \refFigFull{distill} show reduced noise at both near and far focal planes.
We note that this work \textit{focuses on single-view and does not target improved reconstruction fidelity under pupil shift};
the Random \POH results are included only to show that our representation can be used to generate Random \POH for
commercial phase-only holographic display in practice.
Additional experimental results are provided in Suppl~\refSupSec{extra_exp}.

\begin{figure}[thp]
    \centering
    \includegraphics[width=0.99\columnwidth]{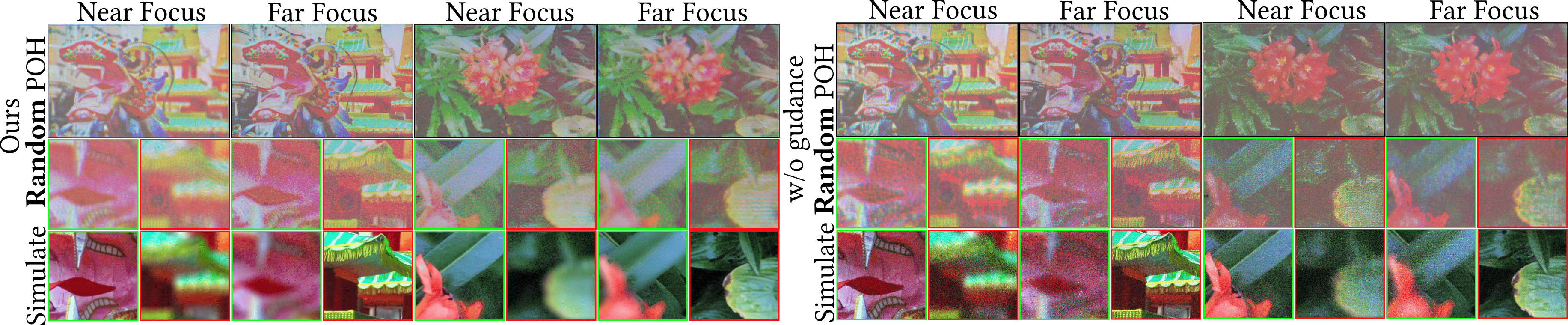}
    \caption{Experimental comparison of reconstructions at near and far focal planes between our method and independently optimized Random \POH (Source Image: ~\cite{dragon2016, mildenhall2019llff}).}
    \label{fig:exp_main}
\end{figure}

\textit{Experiment.} \refFigFull{exp_main} shows experimentally captured results.
While the reduction appears less pronounced than in the simulation due to optical
imperfections and laser speckle, our method still improves visual quality compared with independently optimized Random \POH.

%

\begin{figure}[thp]
    \centering
    \includegraphics[width=0.99\textwidth]{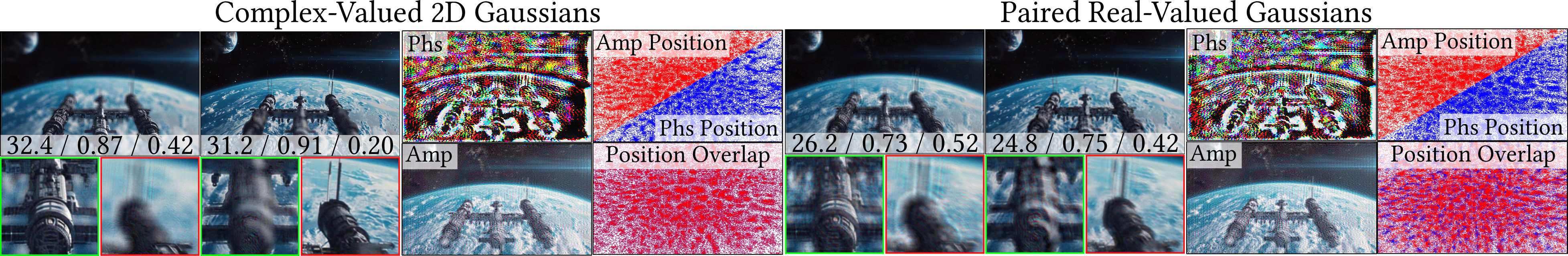}
    \caption{Simulated reconstructions between complex-valued 2D Gaussians and naive paired real-valued Gaussians. (Source:~\cite{spaceship2020})}
    \label{fig:pair_vs_uni}
\end{figure}

\noindent\textbf{Unified Complex vs. Naive Paired Gaussians.}
As shown in \refTbl{ablation_poh}, we compare our unified complex-valued 2D Gaussians against naive paired real-valued 2D Gaussians,
which model amplitude and phase using two independent primitives rendered separately with the standard gsplat~\cite{ye2025gsplat} pipeline.
Our method achieves higher reconstruction quality ($+6.3$ dB) with $33\%$ fewer parameters ($12$ vs. $9\times2$ per pair).
As shown in \refFigFull{pair_vs_uni}, the paired formulation introduces implicit spatial misalignment between amplitude and phase Gaussians,
which distorts the local complex field and leads to blurred reconstructions.
Our unified formulation enforces spatial alignment within each primitive, maintaining sharper details across focal planes.
The paired approach is also an order of magnitude slower due to the overhead of two separate rasterization passes.
These results show that the unified complex-valued Gaussian, as Gabor's elementary signal, is a more effective primitive for hologram representation.

\section{Discussion}
Beyond the core contributions described above, we observe an interesting side effect of our representation in the context of Random \POH conversion.
As shown in \refFigFull{distill} and \refFigFull{exp_main},
our conversion procedure appears to suppress noise artifacts in hologram reconstructions.
We attribute this behavior to the structured nature of our representation:
the complex-valued 2D Gaussians provide a compact, spatially regularized prior to the hologram field,
so the subsequent optimization explores a more constrained parameter search space,
which yields cleaner reconstructions during Random \POH conversion.
However, the current conversion procedure remains indirect,
relying on a separate Gaussian-guided step.
A more elegant approach would be using our representation to directly render Random \POH,
which we consider as a promising future direction.

Our method focuses on \textit{improving reconstruction quality at the center of the eyebox,
whereas existing eyebox-expansion approaches~\cite{choi2022time, chakravarthula2022pupil}
target on complementary objectives rather than serving as direct competitors}.
In Suppl~\refSupSec{eyebox},
we provide an eyebox-shifting results and spectrum analysis for our random \POH.
Additionally, the most pronounced performance gains are observed in pure simulation,
while experimentally captured results show smaller relative margins.
We attribute this gap to several compounding factors: laser speckle introduces multiplicative noise absent in simulation;
$8$-bit \SLM phase quantization and its nonlinear response perturb the encoded field;
and optical misalignment, aberrations, and sensor noise further degrade the capture.
As these factors affect all methods alike, they compress the relative quality margins rather than change the ranking,
which also partly explains the reduced noise-suppression benefit observed on hardware.
Therefore, another promising direction for future research is to extend our method to feedforward, real-time video cases and integrate it with multiplexing
and camera-in-the-loop techniques to further improve the perceptual quality of the reconstructions~\cite{nam2023depolarized, Praneeth2020hitl, choi2022time, Brian2024SigAsia}.

\section{Conclusion}
In this paper, we extend the complex-valued 3D Gaussian formulation introduced for holographic radiance fields~\cite{zhan2025complex}
to the hologram optimization domain.
We propose a novel representation based on complex-valued 2D Gaussian primitives,
each of which acts as Gabor's elementary signal~\cite{gabor1946theory} attaining the minimum space--frequency uncertainty,
supported by a differentiable rasterizer and GPU-optimized light propagation.
Unlike end-to-end \POH methods, our representation serves as a general intermediate convertible to \emph{both} smooth and random \POH,
decoupling the representation from hardware-specific encoding; using it as structural guidance to reduce the search space of the hologram optimization.
Experiments show that our method enables a 5:1 parameter reduction,
50\% faster optimization, and 30\% lower memory consumption,
with reconstruction fidelity comparable to existing methods.

\section*{Acknowledgements}
The authors thank Ye Mao and Dr. Suyeon Choi for the valuable suggestions in the early stage of this work.

\bibliographystyle{splncs04}
\bibliography{main}

\begin{thebibliography}{10}
\providecommand{\url}[1]{\texttt{#1}}
\providecommand{\urlprefix}{URL }
\providecommand{\doi}[1]{https://doi.org/#1}

\bibitem{pomegranate}
Mango, grapefruit, pomegranate, tropical fruit.
  \href{https://openverse.org/image/44348f32-4099-44a1-8f5b-0ad971d51444?q=tropical+fruits&p=35}{Openverse}

\bibitem{aksit2023holobeam}
Ak{\c{s}}it, K., Itoh, Y.: Holobeam: Paper-thin near-eye displays. In: 2023
  IEEE Conference Virtual Reality and 3D User Interfaces (VR). pp. 581--591.
  IEEE (2023)

\bibitem{dragon2016}
Alphab.fr: Chinese new year 2016 in london.
  \href{https://openverse.org/image/da27ffef-2b95-4eb0-b078-1abfcf819f77?q=london&p=1}{Openverse}
  (2016)

\bibitem{AnotherLunch2011}
anotherlunch.com: Soft pretzel, carrots, grapes, cheddar cheese -
  easylunchboxes \& tomica bento.
  \href{https://openverse.org/image/a1c7d020-1d46-4193-8708-a959c5befdb4?q=bento&p=119}{Openverse}
  (2011)

\bibitem{tiger2015}
Appel, M.: Siberian tiger.
  \href{https://openverse.org/image/2ae0e55b-094b-4fa5-9de4-e8682231d1de?q=tiger&p=55}{Openverse}
  (2015)

\bibitem{LAM2021050011}
Blanche, P.A.: Holography, and the future of 3d display. Light: Advanced
  Manufacturing  \textbf{2}(4),  446--459 (2021).
  \doi{https://doi.org/10.37188/lam.2021.028},
  \url{https://www.light-am.com/en/article/doi/10.37188/lam.2021.028}

\bibitem{bohan2023tiny}
Bohan, O.B.: Tiny autoencoder for stable diffusion. Retrieved May  \textbf{22},
  ~2024 (2023)

\bibitem{chakravarthula2022pupil}
Chakravarthula, P., Baek, S.H., Schiffers, F., Tseng, E., Kuo, G., Maimone, A.,
  Matsuda, N., Cossairt, O., Lanman, D., Heide, F.: Pupil-aware holography. ACM
  Trans. Graph.  \textbf{41}(6) (Nov 2022). \doi{10.1145/3550454.3555508},
  \url{https://doi.org/10.1145/3550454.3555508}

\bibitem{Wirtinger2019}
Chakravarthula, P., Peng, Y., Kollin, J., Fuchs, H., Heide, F.: Wirtinger
  holography for near-eye displays. ACM Trans. Graph.  \textbf{38}(6) (Nov
  2019). \doi{10.1145/3355089.3356539},
  \url{https://doi.org/10.1145/3355089.3356539}

\bibitem{Praneeth2020hitl}
Chakravarthula, P., Tseng, E., Srivastava, T., Fuchs, H., Heide, F.: Learned
  hardware-in-the-loop phase retrieval for holographic near-eye displays. ACM
  Trans. Graph.  \textbf{39}(6) (Nov 2020). \doi{10.1145/3414685.3417846},
  \url{https://doi.org/10.1145/3414685.3417846}

\bibitem{Brian2024SigAsia}
Chao, B., Gopakumar, M., Choi, S., Kim, J., Shi, L., Wetzstein, G.: Large
  \'{E}tendue 3d holographic display with content-adaptive dynamic fourier
  modulation. In: SIGGRAPH Asia 2024 Conference Papers. SA '24, Association for
  Computing Machinery, New York, NY, USA (2024). \doi{10.1145/3680528.3687600},
  \url{https://doi.org/10.1145/3680528.3687600}

\bibitem{chao2025random}
Chao, B., Yang, J., Choi, S., Gopakumar, M., Koiso, R., Wetzstein, G.:
  Random-phase wave splatting of translucent primitives for computer-generated
  holography. arXiv preprint arXiv:2508.17480  (2025)

\bibitem{chen2025view}
Chen, K., Wen, A., Zhang, Y., Chakravarthula, P., Sun, Q.: View synthesis for
  3d computer-generated holograms using deep neural fields. Optics Express
  \textbf{33}(9),  19399--19408 (2025)

\bibitem{choi2025GWS}
Choi, S., Chao, B., Yang, J., Gopakumar, M., Wetzstein, G.: Gaussian wave
  splatting for computer-generated holography. ACM Trans. Graph.
  \textbf{44}(4) (Jul 2025). \doi{10.1145/3731163},
  \url{https://doi.org/10.1145/3731163}

\bibitem{choi2022time}
Choi, S., Gopakumar, M., Peng, Y., Kim, J., O'Toole, M., Wetzstein, G.:
  Time-multiplexed neural holography: a flexible framework for holographic
  near-eye displays with fast heavily-quantized spatial light modulators. In:
  ACM SIGGRAPH 2022 Conference Proceedings. pp.~1--9 (2022)

\bibitem{Choi2021Neural3D}
Choi, S., Gopakumar, M., Peng, Y., Kim, J., Wetzstein, G.: Neural 3d
  holography: learning accurate wave propagation models for 3d holographic
  virtual and augmented reality displays. ACM Trans. Graph.  \textbf{40}(6)
  (Dec 2021). \doi{10.1145/3478513.3480542},
  \url{https://doi.org/10.1145/3478513.3480542}

\bibitem{Chu2025RealTime}
Chu, V., Pueyo-Ciutad, O., Tseng, E., Schiffers, F., Kuo, G., Matsuda, N.,
  Redo-Sanchez, A., Lanman, D., Cossairt, O., Heide, F.: Artifact-resilient
  real-time holography. ACM Trans. Graph.  \textbf{44}(6) (Dec 2025).
  \doi{10.1145/3763361}, \url{https://doi.org/10.1145/3763361}

\bibitem{fan2026compressing}
Fan, X., Zhan, Y., Mazumdar, A., Ak{\c{s}}it, K.: {Compressing Double-Phase
  Holograms using 2D Gaussians}. In: Gerrits, T., Teschner, M. (eds.)
  Eurographics 2026 - Posters. The Eurographics Association (2026).
  \doi{10.2312/egp.20261010}

\bibitem{gabor1946theory}
Gabor, D.: Theory of communication. part 1: The analysis of information.
  Journal of the Institution of Electrical Engineers-part III: radio and
  communication engineering  \textbf{93}(26),  429--441 (1946)

\bibitem{gao2025mani}
Gao, X., Li, X., Zhuang, Y., Zhang, Q., Hu, W., Zhang, C., Yao, Y., Shan, Y.,
  Quan, L.: Mani-gs: Gaussian splatting manipulation with triangular mesh. In:
  Proceedings of the Computer Vision and Pattern Recognition Conference. pp.
  21392--21402 (2025)

\bibitem{goodman2005Fourier}
Goodman, J.W.: Introduction to Fourier optics. Roberts and Company publishers
  (2005)

\bibitem{spaceship2020}
HONGSESISHEN: A space station based in a universe inspired by the space opera
  dune by frank herbert. it circles a planet and a big spaceship of the spice
  guild is seen in the background.
  \href{https://openverse.org/image/da315074-5b05-4fc4-b35c-cf69314cda2a?q=blue+planet&p=3}{Openverse}
  (2020)

\bibitem{hossein2020deepcgh}
Hossein~Eybposh, M., Caira, N.W., Atisa, M., Chakravarthula, P., P{\'e}gard,
  N.C.: Deepcgh: 3d computer-generated holography using deep learning. Optics
  Express  \textbf{28}(18),  26636--26650 (2020)

\bibitem{DoublePhase}
Hsueh, C.K., Sawchuk, A.A.: Computer-generated double-phase holograms. Applied
  optics  \textbf{17}(24),  3874--3883 (1978)

\bibitem{huang20242d}
Huang, B., Yu, Z., Chen, A., Geiger, A., Gao, S.: 2d gaussian splatting for
  geometrically accurate radiance fields. In: ACM SIGGRAPH 2024 conference
  papers. pp. 1--11 (2024)

\bibitem{redcar2012}
Hugo-90: 1950s diamond t.
  \href{https://openverse.org/image/f6fd2c82-af63-403f-8e4d-2593f494d04a?q=diamond&p=110}{Openverse}
  (2012)

\bibitem{kavakli2023realistic}
Kavakl{\i}, K., Itoh, Y., Urey, H., Ak{\c{s}}it, K.: Realistic defocus blur for
  multiplane computer-generated holography. In: 2023 IEEE Conference Virtual
  Reality and 3D User Interfaces (VR). pp. 418--426. IEEE (2023)

\bibitem{kavakli2023multicolor}
Kavakl{\i}, K., Shi, L., Urey, H., Matusik, W., Akşit, K.: Multi-color
  holograms improve brightness in holographic displays. In: ACM SIGGRAPH ASIA
  2023 Conference Proceedings. pp.~--. ACM, Sydney, NSW, Australia (2023).
  \doi{https://doi.org/10.1145/3610548.3618135}

\bibitem{Burger2014}
kennejima: Guy fieris vegas kitchen \& bar.
  \href{https://openverse.org/image/4a8a7674-7207-4f15-a8ce-c54abe579c5d?q=burger}{Openverse}
  (2014)

\bibitem{kerbl20233d}
Kerbl, B., Kopanas, G., Leimk{\"u}hler, T., Drettakis, G.: 3d gaussian
  splatting for real-time radiance field rendering. ACM Trans. Graph.
  \textbf{42}(4),  139--1 (2023)

\bibitem{kim2024holographic}
Kim, D., Nam, S.W., Choi, S., Seo, J.M., Wetzstein, G., Jeong, Y.: Holographic
  parallax improves 3d perceptual realism. ACM Transactions on Graphics (TOG)
  \textbf{43}(4),  1--13 (2024)

\bibitem{kuo2023multisource}
Kuo, G., Schiffers, F., Lanman, D., Cossairt, O., Matsuda, N.: Multisource
  holography. ACM Transactions on Graphics (Tog)  \textbf{42}(6),  1--14 (2023)

\bibitem{bluecat2012}
kuujinbo: Manette, wa graffiti wall.
  \href{https://openverse.org/image/772689bc-38b7-44df-8f31-e48292089b31?q=graffiti&p=233}{Openverse}
  (2012)

\bibitem{lindell2022bacon}
Lindell, D.B., Van~Veen, D., Park, J.J., Wetzstein, G.: Bacon: Band-limited
  coordinate networks for multiscale scene representation. In: Proceedings of
  the IEEE/CVF conference on computer vision and pattern recognition. pp.
  16252--16262 (2022)

\bibitem{liu20234k}
Liu, K., Wu, J., He, Z., Cao, L.: 4k-dmdnet: diffraction model-driven network
  for 4k computer-generated holography. Opto-Electronic Advances
  \textbf{6}(5),  220135--1 (2023)

\bibitem{CosineAnnealingLR}
Loshchilov, I., Hutter, F.: {SGDR:} stochastic gradient descent with warm
  restarts. In: 5th International Conference on Learning Representations,
  {ICLR} 2017, Toulon, France, April 24-26, 2017, Conference Track Proceedings.
  OpenReview.net (2017), \url{https://openreview.net/forum?id=Skq89Scxx}

\bibitem{mallick2024taming}
Mallick, S.S., Goel, R., Kerbl, B., Steinberger, M., Carrasco, F.V.,
  De~La~Torre, F.: Taming 3dgs: High-quality radiance fields with limited
  resources. In: SIGGRAPH Asia 2024 Conference Papers. pp. 1--11 (2024).
  \doi{10.1145/3680528.3687694}

\bibitem{matsushima2009band}
Matsushima, K., Shimobaba, T.: Band-limited angular spectrum method for
  numerical simulation of free-space propagation in far and near fields. Optics
  express  \textbf{17}(22),  19662--19673 (2009)

\bibitem{mildenhall2019llff}
Mildenhall, B., Srinivasan, P.P., Ortiz-Cayon, R., Kalantari, N.K.,
  Ramamoorthi, R., Ng, R., Kar, A.: Local light field fusion: Practical view
  synthesis with prescriptive sampling guidelines. ACM Transactions on Graphics
  (TOG)  (2019), \url{https://bmild.github.io/llff/}

\bibitem{nam2023depolarized}
Nam, S.W., Kim, Y., Kim, D., Jeong, Y.: Depolarized holography with
  polarization-multiplexing metasurface. ACM Transactions on Graphics (TOG)
  \textbf{42}(6),  1--16 (2023)

\bibitem{NEURIPS2019_9015}
Paszke, A., et~al.: Pytorch: An imperative style, high-performance deep
  learning library. In: Advances in Neural Information Processing Systems 32,
  pp. 8024--8035. Curran Associates, Inc. (2019),
  \url{http://papers.neurips.cc}

\bibitem{peng2020neural}
Peng, Y., Choi, S., Padmanaban, N., Wetzstein, G.: Neural holography with
  camera-in-the-loop training. ACM Transactions on Graphics (TOG)
  \textbf{39}(6),  1--14 (2020)

\bibitem{Peng2025Poster}
Peng, Z., Zhan, Y., Spjut, J., Ak\c{s}it, K.: Assessing learned models for
  phase-only hologram compression. In: Proceedings of the Special Interest
  Group on Computer Graphics and Interactive Techniques Conference Posters.
  SIGGRAPH Posters '25, Association for Computing Machinery, New York, NY, USA
  (2025). \doi{10.1145/3721250.3742993},
  \url{https://doi.org/10.1145/3721250.3742993}

\bibitem{preechakul2022diffusion}
Preechakul, K., Chatthee, N., Wizadwongsa, S., Suwajanakorn, S.: Diffusion
  autoencoders: Toward a meaningful and decodable representation. In:
  Proceedings of the IEEE/CVF conference on computer vision and pattern
  recognition. pp. 10619--10629 (2022)

\bibitem{Ranftl2022}
Ranftl, R., Lasinger, K., Hafner, D., Schindler, K., Koltun, V.: Towards robust
  monocular depth estimation: Mixing datasets for zero-shot cross-dataset
  transfer. IEEE Transactions on Pattern Analysis and Machine Intelligence
  \textbf{44}(3) (2022)

\bibitem{ronneberger2015u}
Ronneberger, O., Fischer, P., Brox, T.: U-net: Convolutional networks for
  biomedical image segmentation. In: Medical image computing and
  computer-assisted intervention--MICCAI 2015: 18th international conference,
  Munich, Germany, October 5-9, 2015, proceedings, part III 18. pp. 234--241.
  Springer (2015)

\bibitem{rumelhart1986learning}
Rumelhart, D.E., Hinton, G.E., Williams, R.J.: Learning representations by
  back-propagating errors. nature  \textbf{323}(6088),  533--536 (1986)

\bibitem{saragadam2023wire}
Saragadam, V., LeJeune, D., Tan, J., Balakrishnan, G., Veeraraghavan, A.,
  Baraniuk, R.G.: Wire: Wavelet implicit neural representations. In:
  Proceedings of the IEEE/CVF Conference on Computer Vision and Pattern
  Recognition. pp. 18507--18516 (2023)

\bibitem{schiffers2023stochastic}
Schiffers, F., Chakravarthula, P., Matsuda, N., Kuo, G., Tseng, E., Lanman, D.,
  Heide, F., Cossairt, O.: Stochastic light field holography. IEEE
  International Conference on Computational Photography (ICCP)  (2023)

\bibitem{Schiffers2025multiwavelength}
Schiffers, F.A., Kuo, G., Matsuda, N., Lanman, D., Cossairt, O.: Holochrome:
  Polychromatic illumination for speckle reduction in holographic near-eye
  displays. ACM Trans. Graph.  \textbf{44}(3) (May 2025).
  \doi{10.1145/3732935}, \url{https://doi.org/10.1145/3732935}

\bibitem{10.1145/3130800.3130832}
Shi, L., Huang, F.C., Lopes, W., Matusik, W., Luebke, D.: Near-eye light field
  holographic rendering with spherical waves for wide field of view interactive
  3d computer graphics. ACM Trans. Graph.  \textbf{36}(6) (Nov 2017).
  \doi{10.1145/3130800.3130832}, \url{https://doi.org/10.1145/3130800.3130832}

\bibitem{shi2021towards}
Shi, L., Li, B., Kim, C., Kellnhofer, P., Matusik, W.: Towards real-time
  photorealistic 3d holography with deep neural networks. Nature
  \textbf{591}(7849),  234--239 (2021)

\bibitem{shi2022end}
Shi, L., Li, B., Matusik, W.: End-to-end learning of 3d phase-only holograms
  for holographic display. Light: Science \& Applications  \textbf{11}(1), ~247
  (2022)

\bibitem{Shi2022OL}
Shi, L., Webb, R., Xiao, L., Kim, C., Jang, C.: Neural compression for hologram
  images and videos. Opt. Lett.  \textbf{47}(22), ~6013 (Nov 2022)

\bibitem{sitzmann2020implicit}
Sitzmann, V., Martel, J., Bergman, A., Lindell, D., Wetzstein, G.: Implicit
  neural representations with periodic activation functions. Advances in neural
  information processing systems  \textbf{33},  7462--7473 (2020)

\bibitem{song2023consistency}
Song, Y., Dhariwal, P., Chen, M., Sutskever, I.: Consistency models. ICML
  (2023)

\bibitem{Sutherland2010}
Sutherland, B.: Wasabi rainbow sushi set.
  \href{https://openverse.org/image/90ca6d2b-8c24-48d5-abf2-6f0dc2d81808?q=sushi&p=10}{Openverse}
  (2010)

\bibitem{Wilson2009}
over 10 million~views Thanks~!!, S.W.: A pair of ecuadorian amazon red-lored
  parrots.
  \href{https://openverse.org/image/6ee600ad-2f1e-4df9-957d-4b1efcb7e2c4?q=colorful+city&p=123}{Openverse}
  (2009)

\bibitem{Timofte_2017_CVPR_Workshops}
Timofte, R., Agustsson, E., Van~Gool, L., Yang, M.H., Zhang, L., Lim, B.,
  et~al.: Ntire 2017 challenge on single image super-resolution: Methods and
  results. In: The IEEE Conference on Computer Vision and Pattern Recognition
  (CVPR) Workshops (July 2017)

\bibitem{wang2024sinsr}
Wang, Y., Yang, W., Chen, X., Wang, Y., Guo, L., Chau, L.P., Liu, Z., Qiao, Y.,
  Kot, A.C., Wen, B.: Sinsr: diffusion-based image super-resolution in a single
  step. In: Proceedings of the IEEE/CVF conference on computer vision and
  pattern recognition. pp. 25796--25805 (2024)

\bibitem{wang2022joint}
Wang, Y., Chakravarthula, P., Sun, Q., Chen, B.: Joint neural phase retrieval
  and compression for energy-and computation-efficient holography on the edge.
  ACM Transactions on Graphics  \textbf{41}(4) (2022)

\bibitem{wu20244d}
Wu, G., Yi, T., Fang, J., Xie, L., Zhang, X., Wei, W., Liu, W., Tian, Q., Wang,
  X.: 4d gaussian splatting for real-time dynamic scene rendering. In:
  Proceedings of the IEEE/CVF Conference on Computer Vision and Pattern
  Recognition. pp. 20310--20320 (2024)

\bibitem{straw2013}
www.metaphoricalplatypus.com: Peas and strawberries.
  \href{https://openverse.org/image/31e24388-74bd-4ec8-93c7-470e3d2a01e1?q=strawberry\&p=59}{Openverse}
  (2013)

\bibitem{xie2024adan}
Xie, X., Zhou, P., Li, H., Lin, Z., Yan, S.: Adan: Adaptive nesterov momentum
  algorithm for faster optimizing deep models. IEEE Transactions on Pattern
  Analysis and Machine Intelligence  (2024)

\bibitem{depth_anything_v2}
Yang, L., Kang, B., Huang, Z., Zhao, Z., Xu, X., Feng, J., Zhao, H.: Depth
  anything v2. arXiv:2406.09414  (2024)

\bibitem{ye2025gsplat}
Ye, V., Li, R., Kerr, J., Turkulainen, M., Yi, B., Pan, Z., Seiskari, O., Ye,
  J., Hu, J., Tancik, M., et~al.: gsplat: An open-source library for gaussian
  splatting. Journal of Machine Learning Research  \textbf{26}(34),  1--17
  (2025)

\bibitem{Zeng_2025_ICCV}
Zeng, Z., Wang, Y., Yang, C., Guan, T., Ju, L.: Instant gaussianimage: A
  generalizable and self-adaptive image representation via 2d gaussian
  splatting. In: Proceedings of the IEEE/CVF International Conference on
  Computer Vision (ICCV) (October 2025)

\bibitem{ourcode2026}
Zhan, Y., Gao, X., Quan, L., Ak{\c{s}}it, K.: Complex-valued 2d gaussian
  representation for computer-generated holography.
  \url{https://github.com/complight/Complex-Valued_2D_Gaussian_Representation}
  (2026), source code

\bibitem{zhan2025complex}
Zhan, Y., Shin, D.H., Baek, S.H., Ak{\c{s}}it, K.: Complex-valued holographic
  radiance fields. ACM Trans. Graph.  \textbf{45}(3) (Apr 2026).
  \doi{10.1145/3804450}, \url{https://doi.org/10.1145/3804450}

\bibitem{zhan2024Configure}
Zhan, Y., Sun, Q., Shi, L., Matusik, W., Akşit, K.: Configurable holography:
  Towards display and scene adaptation. arXiv preprint arXiv:2405.01558  (2024)

\bibitem{zhang2019LPIPS}
Zhang, R., Isola, P., Efros, A.A., Shechtman, E., Wang, O.: The unreasonable
  effectiveness of deep features as a perceptual metric. In: 2018 {IEEE}
  Conference on Computer Vision and Pattern Recognition, {CVPR} 2018, Salt Lake
  City, UT, USA, June 18-22, 2018. pp. 586--595. Computer Vision Foundation /
  {IEEE} Computer Society (2018). \doi{10.1109/CVPR.2018.00068},
  \url{http://openaccess.thecvf.com/content\_cvpr\_2018/html/Zhang\_The\_Unreasonable\_Effectiveness\_CVPR\_2018\_paper.html}

\bibitem{zhang2024gaussianimage}
Zhang, X., Ge, X., Xu, T., He, D., Wang, Y., Qin, H., Lu, G., Geng, J., Zhang,
  J.: Gaussianimage: 1000 fps image representation and compression by 2d
  gaussian splatting. In: European Conference on Computer Vision. pp. 327--345.
  Springer (2024)

\bibitem{zhang2025image}
Zhang, Y., Li, B., Kuznetsov, A., Jindal, A., Diolatzis, S., Chen, K.,
  Sochenov, A., Kaplanyan, A., Sun, Q.: Image-gs: Content-adaptive image
  representation via 2d gaussians. In: Proceedings of the Special Interest
  Group on Computer Graphics and Interactive Techniques Conference Conference
  Papers. pp. 1--11 (2025)

\bibitem{Chuanjun2024SigAsia}
Zheng, C., Zhan, Y., Shi, L., Cakmakci, O., Ak\c{s}it, K.: Focal surface
  holographic light transport using learned spatially adaptive convolutions.
  In: SIGGRAPH Asia 2024 Technical Communications. SA '24, Association for
  Computing Machinery, New York, NY, USA (2024). \doi{10.1145/3681758.3697989},
  \url{https://doi.org/10.1145/3681758.3697989}

\end{thebibliography}


\begin{thebibliography}{10}
\providecommand{\url}[1]{\texttt{#1}}
\providecommand{\urlprefix}{URL }
\providecommand{\doi}[1]{https://doi.org/#1}

\bibitem{dragon2016}
Alphab.fr: Chinese new year 2016 in london.
  \href{https://openverse.org/image/da27ffef-2b95-4eb0-b078-1abfcf819f77?q=london&p=1}{Openverse}
  (2016)

\bibitem{AnotherLunch2011}
anotherlunch.com: Soft pretzel, carrots, grapes, cheddar cheese -
  easylunchboxes \& tomica bento.
  \href{https://openverse.org/image/a1c7d020-1d46-4193-8708-a959c5befdb4?q=bento&p=119}{Openverse}
  (2011)

\bibitem{tiger2015}
Appel, M.: Siberian tiger.
  \href{https://openverse.org/image/2ae0e55b-094b-4fa5-9de4-e8682231d1de?q=tiger&p=55}{Openverse}
  (2015)

\bibitem{chakravarthula2022pupil}
Chakravarthula, P., Baek, S.H., Schiffers, F., Tseng, E., Kuo, G., Maimone, A.,
  Matsuda, N., Cossairt, O., Lanman, D., Heide, F.: Pupil-aware holography. ACM
  Trans. Graph.  \textbf{41}(6) (Nov 2022). \doi{10.1145/3550454.3555508},
  \url{https://doi.org/10.1145/3550454.3555508}

\bibitem{choi2022time}
Choi, S., Gopakumar, M., Peng, Y., Kim, J., O'Toole, M., Wetzstein, G.:
  Time-multiplexed neural holography: a flexible framework for holographic
  near-eye displays with fast heavily-quantized spatial light modulators. In:
  ACM SIGGRAPH 2022 Conference Proceedings. pp.~1--9 (2022)

\bibitem{Choi2021Neural3D}
Choi, S., Gopakumar, M., Peng, Y., Kim, J., Wetzstein, G.: Neural 3d
  holography: learning accurate wave propagation models for 3d holographic
  virtual and augmented reality displays. ACM Trans. Graph.  \textbf{40}(6)
  (Dec 2021). \doi{10.1145/3478513.3480542},
  \url{https://doi.org/10.1145/3478513.3480542}

\bibitem{Chu2025RealTime}
Chu, V., Pueyo-Ciutad, O., Tseng, E., Schiffers, F., Kuo, G., Matsuda, N.,
  Redo-Sanchez, A., Lanman, D., Cossairt, O., Heide, F.: Artifact-resilient
  real-time holography. ACM Trans. Graph.  \textbf{44}(6) (Dec 2025).
  \doi{10.1145/3763361}, \url{https://doi.org/10.1145/3763361}

\bibitem{redcar2012}
Hugo-90: 1950s diamond t.
  \href{https://openverse.org/image/f6fd2c82-af63-403f-8e4d-2593f494d04a?q=diamond&p=110}{Openverse}
  (2012)

\bibitem{kavakli2023multicolor}
Kavakl{\i}, K., Shi, L., Urey, H., Matusik, W., Akşit, K.: Multi-color
  holograms improve brightness in holographic displays. In: ACM SIGGRAPH ASIA
  2023 Conference Proceedings. pp.~--. ACM, Sydney, NSW, Australia (2023).
  \doi{https://doi.org/10.1145/3610548.3618135}

\bibitem{kim2024holographic}
Kim, D., Nam, S.W., Choi, S., Seo, J.M., Wetzstein, G., Jeong, Y.: Holographic
  parallax improves 3d perceptual realism. ACM Transactions on Graphics (TOG)
  \textbf{43}(4),  1--13 (2024)

\bibitem{kuo2023multisource}
Kuo, G., Schiffers, F., Lanman, D., Cossairt, O., Matsuda, N.: Multisource
  holography. ACM Transactions on Graphics (Tog)  \textbf{42}(6),  1--14 (2023)

\bibitem{mildenhall2019llff}
Mildenhall, B., Srinivasan, P.P., Ortiz-Cayon, R., Kalantari, N.K.,
  Ramamoorthi, R., Ng, R., Kar, A.: Local light field fusion: Practical view
  synthesis with prescriptive sampling guidelines. ACM Transactions on Graphics
  (TOG)  (2019), \url{https://bmild.github.io/llff/}

\bibitem{schiffers2023stochastic}
Schiffers, F., Chakravarthula, P., Matsuda, N., Kuo, G., Tseng, E., Lanman, D.,
  Heide, F., Cossairt, O.: Stochastic light field holography. IEEE
  International Conference on Computational Photography (ICCP)  (2023)

\bibitem{Schiffers2025multiwavelength}
Schiffers, F.A., Kuo, G., Matsuda, N., Lanman, D., Cossairt, O.: Holochrome:
  Polychromatic illumination for speckle reduction in holographic near-eye
  displays. ACM Trans. Graph.  \textbf{44}(3) (May 2025).
  \doi{10.1145/3732935}, \url{https://doi.org/10.1145/3732935}

\bibitem{straw2013}
www.metaphoricalplatypus.com: Peas and strawberries.
  \href{https://openverse.org/image/31e24388-74bd-4ec8-93c7-470e3d2a01e1?q=strawberry\&p=59}{Openverse}
  (2013)

\bibitem{zhang2024gaussianimage}
Zhang, X., Ge, X., Xu, T., He, D., Wang, Y., Qin, H., Lu, G., Geng, J., Zhang,
  J.: Gaussianimage: 1000 fps image representation and compression by 2d
  gaussian splatting. In: European Conference on Computer Vision. pp. 327--345.
  Springer (2024)

\end{thebibliography}

\clearpage
\label{supp:firstpage}
\begin{center}
{\LARGE\bfseries Supplementary Material: Complex-Valued 2D Gaussian Representation for Computer-Generated Holography}
\vspace{1em}
\end{center}

\section{Hardware Prototype}
\label{supplementary:hardware}

\begin{figure}[ht!]
  \centering
  \includegraphics[width=0.45\textwidth]{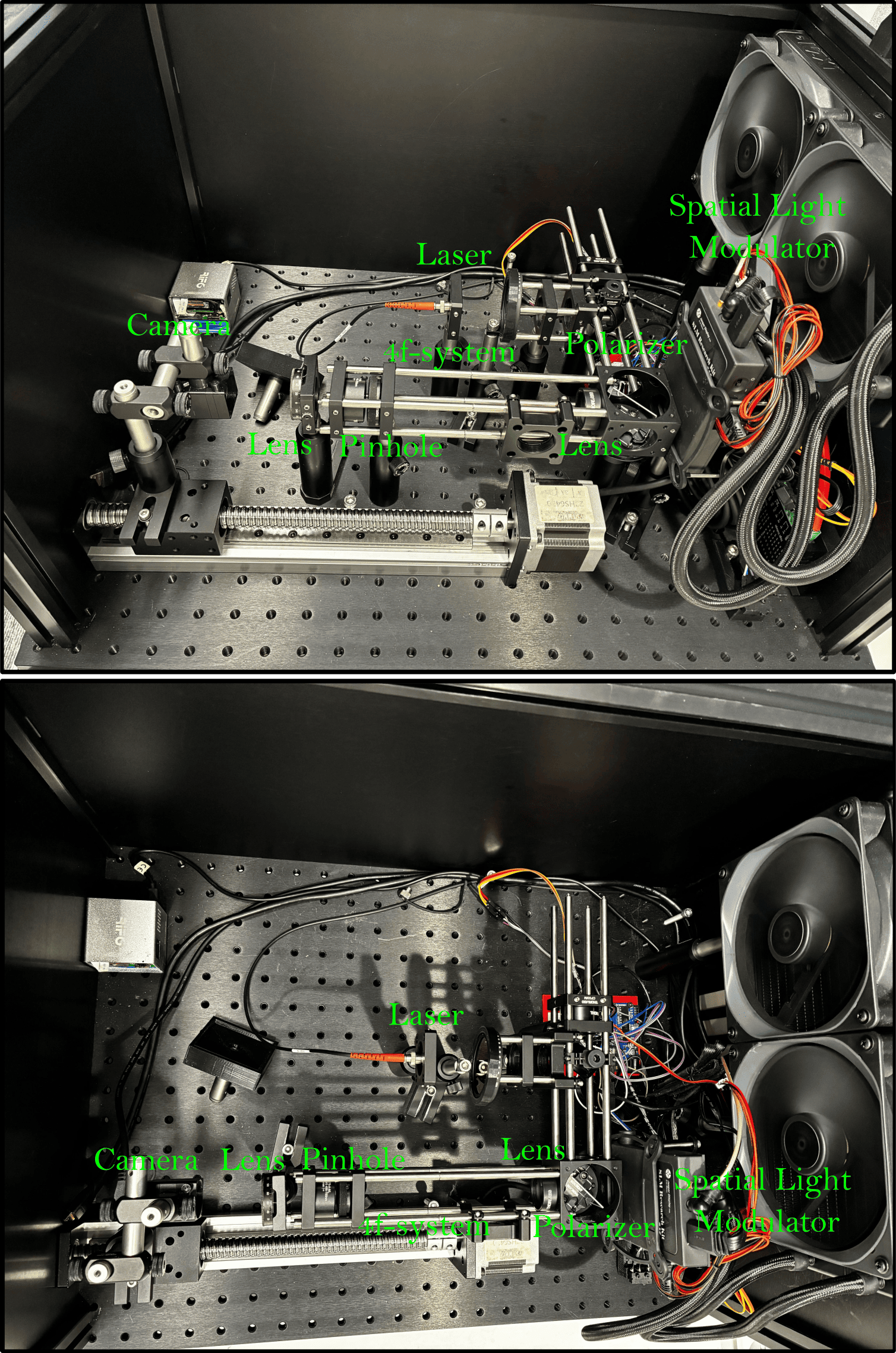}
  \caption{Holographic display prototype (Jasper JD7714) used to evaluate holograms generated by our model.}
  \label{fig:hardware_jasper}
\end{figure}

Figure~\ref{fig:hardware_jasper} shows the holographic display prototype used in our experiments.
The optical path begins with a laser source (LASOS MCS4) integrating three individual laser lines.
Light emitted from a single-mode fibre is collimated by a Thorlabs LA1708-A plano-convex lens with a 200~mm focal length.
The linearly polarized, collimated beam is then directed by a beamsplitter (Thorlabs BP245B1) onto a phase-only \SLM, the Jasper JD7714 ($2400 \times 4094$, $3.74~\mu$m pixel pitch).
The modulated beam passes through a lens pair (Thorlabs LA1908-A and LB1056-A) with focal lengths of 500~mm and 250~mm, respectively.
A pinhole aperture (Thorlabs SM1D12) is placed at the focal plane of the lenses for spatial filtering.
Finally, the holographic reconstructions are recorded by a lensless image sensor (Point Grey GS3-U3-23S6M-C, USB 3.0) mounted on an X-stage (Thorlabs PT1/M) with a travel range of 0--25~mm and a positioning precision of 0.01~mm.

\section{Differentiable 2D Complex-Valued Gaussian Rendering}
\label{supplementary:2d_cuda_gradient}

This section provides detailed mathematical formulations and gradient derivations for our 2D Complex-Valued Gaussian Rasterizer.

\subsection{Notation}
\begin{itemize}
  \item $s_x, s_y \in \mathbb{R}^+$ - Activated scaling factors for Gaussian in $x$ and $y$ directions
  \item $\tilde{s}_x, \tilde{s}_y \in \mathbb{R}$ - Pre-activation scale parameters
  \item $\theta \in \mathbb{R}$ - Rotation angle of the Gaussian ellipse
  \item $\mathbf{x} = (x_0, x_1) \in \mathbb{R}^2$ - Activated 2D mean position
  \item $\tilde{\mathbf{x}} = (\tilde{x}_0, \tilde{x}_1) \in \mathbb{R}^2$ - Pre-activation mean parameters
  \item $\Sigma \in \mathbb{R}^{2 \times 2}$ - 2D covariance matrix
  \item $\Sigma^{-1} \in \mathbb{R}^{2 \times 2}$ - Inverse 2D covariance matrix
  \item $\Sigma^{-1}_{ij}$ - Elements of inverse covariance where $i,j \in \{0,1\}$
  \item $d_x = x - x_0, d_y = y - x_1$ - Distance from pixel to Gaussian center
  \item $\mathbf{c}_{n} \in \mathbb{R}^C$ - Color/amplitude values for Gaussian $n$ across $C$ channels
  \item $\boldsymbol{\varphi}_{n} \in \mathbb{R}^C$ - Phase values for Gaussian $n$ across $C$ channels
  \item $\alpha_n \in [0,1]$ - Activated opacity value for Gaussian $n$
  \item $\tilde{\alpha}_n \in \mathbb{R}$ - Pre-activation opacity parameter for Gaussian $n$
  \item $\text{power}$ - Gaussian exponent term (negative half Mahalanobis distance)
  \item $W, H$ - Image width and height
  \item $\epsilon_s = 0.1$ - Scale regularization constant
  \item $\epsilon_c = 0.1$ - Covariance regularization constant
  \item $\epsilon_d = 10^{-10}$ - Determinant clamping threshold
\end{itemize}

\subsection{Optimization Algorithm}

Our optimization algorithm adapts 2D Gaussian primitives for hologram generation, as summarized in Algorithm~\ref{alg:2d_optimization}.

\begin{algorithm}[ht!]
\caption{2D Gaussian Hologram Optimization}
\label{alg:2d_optimization}
\footnotesize
\begin{algorithmic}[1]
\Require $W, H$: hologram resolution
\Require $L$: number of depth planes
\State $M \leftarrow$ InitPositions($N$)
\State $\mathbf{S}, \mathbf{C}, \alpha \leftarrow$ InitAttributes()
\State $\boldsymbol{\varphi} \leftarrow$ InitPhase()
\State $\theta \leftarrow$ InitRotation()
\State $i \leftarrow 0$
\While{not converged}
    \State $I_{\text{target}}, D \leftarrow$ GetTarget()
    \State $U_{\text{complex}} \leftarrow$ ComplexRasterize2D($M, \mathbf{S},$
    \Statex \hspace{3em} $\theta, \mathbf{C}, \boldsymbol{\varphi}, \alpha$)
    \State $P \leftarrow$ ZeroPad($U_{\text{complex}}$)
    \State $\{I_{\text{recon},l}\}_{l=1}^L \leftarrow$ MultiPlanePropagate($P$)
    \State $\mathcal{L} \leftarrow$ MultiPlaneLoss($\{I_{\text{recon},l}\},$
    \Statex \hspace{3em} $I_{\text{target}}, D$)
    \State $M, \mathbf{S}, \theta, \mathbf{C}, \boldsymbol{\varphi}, \alpha \leftarrow$ Adan($\nabla\mathcal{L}$)
    \State $i \leftarrow i + 1$
\EndWhile
\end{algorithmic}
\end{algorithm}

\subsection{Tile-Based Rasterizer}

Our tile-based rasterizer efficiently computes complex fields across the hologram plane, as detailed in Algorithm~\ref{alg:2d_rasterizer}.

\begin{algorithm}[ht!]
\caption{2D Complex-Valued Tile-Based Rasterization}
\label{alg:2d_rasterizer}
\footnotesize
\begin{algorithmic}[1]
\Require $W, H$: hologram dimensions
\Require $M, \mathbf{S}, \theta$: Gaussian positions, scales, rotations
\Require $\mathbf{C}, \alpha, \boldsymbol{\varphi}$: Amplitudes, opacities, phases
\Require $C$: number of color channels

\Function{ComplexRasterize2D}{$W, H, M, \mathbf{S}, \theta,$}
\Statex \hspace{8em} $\mathbf{C}, \alpha, \boldsymbol{\varphi}$
    \State $\Sigma \leftarrow$ Compute2DCovariance($\mathbf{S}, \theta$)
    \State $\Sigma^{-1}, r \leftarrow$ Invert2DCovariance($\Sigma$)
    \State $T \leftarrow$ CreateTiles($W, H$)
    \State $\mathcal{I}, \mathcal{K} \leftarrow$ DuplicateWithKeys($M, r, T$)
    \State $\mathcal{K}_s, \mathcal{I}_s \leftarrow$ SortByKeys($\mathcal{I}, \mathcal{K}$)
    \State $\mathcal{R} \leftarrow$ IdentifyTileRanges($T, \mathcal{K}_s$)
    \State $U_{\text{real}}, U_{\text{imag}} \leftarrow$ InitCanvas($C, W, H$)
    \ForAll{Tiles $t$ \textbf{in} $T$ \textbf{parallel}}
        \ForAll{Pixels $pix$ \textbf{in} $t$ \textbf{parallel}}
            \State $\text{real}_{\text{acc}}[C], \text{imag}_{\text{acc}}[C] \leftarrow 0$
            \State range $\leftarrow$ GetTileRange($\mathcal{R}, t$)
            \For{$g$ \textbf{in} range}
                \State $d_x \leftarrow pix_x - x_{g,0}$, $d_y \leftarrow pix_y - x_{g,1}$
                \State $\text{power} \leftarrow -0.5 \cdot (d_x^2 \Sigma^{-1}_{00} +$
                \Statex \hspace{10em} $2d_xd_y\Sigma^{-1}_{01} + d_y^2 \Sigma^{-1}_{11})$
                \State $G \leftarrow \exp(\max(\text{power}, -50))$
                \State $\alpha_{\text{eff}} \leftarrow \min(0.99, \alpha_g \cdot G)$
                \If{$\alpha_{\text{eff}} < 1/255$}
                    \State \textbf{continue}
                \EndIf
                \For{$c \leftarrow 0$ \textbf{to} $C-1$}
                    \State $\text{scale} \leftarrow \mathbf{c}_{g,c} \cdot \alpha_{\text{eff}}$
                    \State $\cos_\varphi, \sin_\varphi \leftarrow \cos(\boldsymbol{\varphi}_{g,c}),$
                    \Statex \hspace{12em} $\sin(\boldsymbol{\varphi}_{g,c})$
                    \State $\text{real}_{\text{acc}}[c] \mathrel{+}= \text{scale} \cdot \cos_\varphi$
                    \State $\text{imag}_{\text{acc}}[c] \mathrel{+}= \text{scale} \cdot \sin_\varphi$
                \EndFor
            \EndFor
            \For{$c \leftarrow 0$ \textbf{to} $C-1$}
                \State $U_{\text{real}}[c, pix] \leftarrow \text{real}_{\text{acc}}[c]$
                \State $U_{\text{imag}}[c, pix] \leftarrow \text{imag}_{\text{acc}}[c]$
            \EndFor
        \EndFor
    \EndFor
    \State \Return $U_{\text{real}} + j \cdot U_{\text{imag}}$
\EndFunction
\end{algorithmic}
\end{algorithm}

Key features include: (1) parallel tile processing with $16 \times 16$ blocks, (2) shared VRAM for batch Gaussian loading,
(3) early termination when Gaussian contribution is negligible (e.g., $\alpha_{\text{eff}} < 1/255$),
(4) channel-wise complex accumulation, and (5) improved numerical stability via power clamping.

\subsection{Forward Pass}

\subsubsection{Parameter Activation Functions}
\label{supplementary:activation}

\textbf{Mean Position Activation (Tanh-based):}
\begin{equation}
  \mathbf{x} = \left(\frac{\tanh(\tilde{x}_x) + 1}{2} \cdot W, \frac{\tanh(\tilde{x}_y) + 1}{2} \cdot H\right)
\end{equation}
\textbf{Scale Activation (Exponential):}
\begin{equation}
  s_x = \exp(\tilde{s}_x) + \epsilon_s, \quad s_y = \exp(\tilde{s}_y) + \epsilon_s
\end{equation}
\textbf{Opacity Activation (Sigmoid):}
\begin{equation}
  \alpha_n = \sigma(\tilde{\alpha}_n) = \frac{1}{1 + \exp(-\tilde{\alpha}_n)}
\end{equation}

\subsubsection{2D Covariance Matrix Computation}
\label{supplementary:2d_cov}
The 2D covariance matrix is:
\begin{equation}
  \Sigma = R \cdot S^2 \cdot R^T + \epsilon_c \cdot \mathbf{I}
\end{equation}
where $R = \begin{pmatrix} \cos\theta & -\sin\theta \\ \sin\theta & \cos\theta \end{pmatrix}$ and $S^2 = \begin{pmatrix} s_x^2 & 0 \\ 0 & s_y^2 \end{pmatrix}$.
Expanding:
\begin{equation}
  \begin{aligned}
  \Sigma_{00} &= s_x^2 \cos^2\theta + s_y^2 \sin^2\theta + \epsilon_c \\
  \Sigma_{01} &= (s_x^2 - s_y^2) \cos\theta \sin\theta \\
  \Sigma_{11} &= s_x^2 \sin^2\theta + s_y^2 \cos^2\theta + \epsilon_c
  \end{aligned}
\end{equation}

\subsubsection{2D Covariance Matrix Inversion}

For 2×2 matrix inversion:
\begin{equation}
  \det(\Sigma) = \Sigma_{00} \Sigma_{11} - \Sigma_{01}^2
\end{equation}
\begin{equation}
  \Sigma^{-1} = \frac{1}{\max(\det(\Sigma), \epsilon_d)} \begin{pmatrix} \Sigma_{11} & -\Sigma_{01} \\ -\Sigma_{01} & \Sigma_{00} \end{pmatrix}
\end{equation}
Stored as $[\text{inv}_{00}, \text{inv}_{01}, \text{inv}_{11}]$:
\begin{equation}
  \begin{aligned}
    \text{inv}_{00} &= \Sigma_{11} / \det_{\text{safe}} \\
    \text{inv}_{01} &= -\Sigma_{01} / \det_{\text{safe}} \\
    \text{inv}_{11} &= \Sigma_{00} / \det_{\text{safe}}
  \end{aligned}
\end{equation}

\subsubsection{Gaussian Evaluation}

For pixel $(x,y)$ and Gaussian $n$:
\begin{equation}
  \begin{aligned}
  \text{mahal\_dist} &= d_x^2 \cdot \text{inv}_{00} + 2 d_x d_y \cdot \text{inv}_{01} \\
  &\quad + d_y^2 \cdot \text{inv}_{11}
  \end{aligned}
\end{equation}
where $d_x = x - x_x$, $d_y = y - x_y$.
With numerical stability:
\begin{equation}
  \text{power} = \max(-0.5 \cdot \text{mahal\_dist}, -50.0)
\end{equation}
\begin{equation}
  \mathcal{G}_n(x,y) = \exp(\text{power})
\end{equation}

\subsubsection{Complex Field Rendering}

For each channel $c$:
\begin{equation}
  \begin{aligned}
    \text{real}_c(x,y) &= \sum_n \mathbf{c}_{n,c} \cdot \alpha_n \cdot \mathcal{G}_n(x,y) \cdot \cos(\boldsymbol{\varphi}_{n,c}) \\
    \text{imag}_c(x,y) &= \sum_n \mathbf{c}_{n,c} \cdot \alpha_n \cdot \mathcal{G}_n(x,y) \cdot \sin(\boldsymbol{\varphi}_{n,c})
  \end{aligned}
\end{equation}
\begin{equation}
  U_c(x,y) = \text{real}_c(x,y) + j \cdot \text{imag}_c(x,y)
\end{equation}

\subsection{Backward Pass}

\subsubsection{Gradient Flow Overview}

For parameter $\tilde{\theta}$:
\begin{equation}
  \frac{\partial \mathcal{L}}{\partial \tilde{\theta}} = \frac{\partial \mathcal{L}}{\partial U} \cdot \frac{\partial U}{\partial \theta} \cdot \frac{\partial \theta}{\partial \tilde{\theta}}
\end{equation}
Real and imaginary components are accumulated separately but remain coupled through shared Gaussian parameters.

\subsubsection{Detailed Gradient Derivation}

\begin{enumerate}
\item \textbf{Gradient for Color/Amplitude $\mathbf{c}_n$}

For each channel $c$ and Gaussian $n$:
\begin{equation}
  \begin{aligned}
  \frac{\partial \mathcal{L}}{\partial \mathbf{c}_{n,c}} &= \sum_{x,y} \alpha_n \cdot \mathcal{G}_n(x,y) \cdot \Big(\cos(\boldsymbol{\varphi}_{n,c}) \cdot \frac{\partial \mathcal{L}}{\partial \text{real}_c(x,y)} \\
  &\quad + \sin(\boldsymbol{\varphi}_{n,c}) \cdot \frac{\partial \mathcal{L}}{\partial \text{imag}_c(x,y)}\Big)
  \end{aligned}
\end{equation}
where $\mathcal{G}_n(x,y) = \exp(\text{power})$ with clamping applied.

\item \textbf{Gradient for Phase $\boldsymbol{\varphi}_n$}

\begin{equation}
  \begin{aligned}
  \frac{\partial \mathcal{L}}{\partial \boldsymbol{\varphi}_{n,c}} &= \sum_{x,y} \mathbf{c}_{n,c} \cdot \alpha_n \cdot \mathcal{G}_n(x,y) \cdot \Big(-\sin(\boldsymbol{\varphi}_{n,c}) \\
  &\quad \cdot \frac{\partial \mathcal{L}}{\partial \text{real}_c(x,y)} + \cos(\boldsymbol{\varphi}_{n,c}) \cdot \frac{\partial \mathcal{L}}{\partial \text{imag}_c(x,y)}\Big)
  \end{aligned}
\end{equation}

\item \textbf{Gradient for Pre-activation Opacity $\tilde{\alpha}_n$}

First compute gradient w.r.t. activated opacity:
\begin{equation}
  \begin{aligned}
  \frac{\partial \mathcal{L}}{\partial \alpha_n} &= \sum_{c,x,y} \mathbf{c}_{n,c} \cdot \mathcal{G}_n(x,y) \cdot \Big(\cos(\boldsymbol{\varphi}_{n,c}) \\
  &\quad \cdot \frac{\partial \mathcal{L}}{\partial \text{real}_c(x,y)} + \sin(\boldsymbol{\varphi}_{n,c}) \cdot \frac{\partial \mathcal{L}}{\partial \text{imag}_c(x,y)}\Big)
  \end{aligned}
\end{equation}
Then apply sigmoid derivative:
\begin{equation}
  \frac{\partial \mathcal{L}}{\partial \tilde{\alpha}_n} = \frac{\partial \mathcal{L}}{\partial \alpha_n} \cdot \alpha_n \cdot (1 - \alpha_n)
\end{equation}

\item \textbf{Gradient for Pre-activation Mean $\tilde{\mathbf{x}}_n$}

The gradient flows through: $\tilde{\mathbf{x}} \rightarrow \mathbf{x} \rightarrow d_x, d_y \rightarrow \text{mahal\_dist} \rightarrow \text{power} \rightarrow \mathcal{G}_n$.
For the Mahalanobis distance:
\begin{equation}
  \begin{aligned}
    \frac{\partial \text{mahal\_dist}}{\partial x_x} &= -2(d_x \cdot \text{inv}_{00} + d_y \cdot \text{inv}_{01}) \\
    \frac{\partial \text{mahal\_dist}}{\partial x_y} &= -2(d_x \cdot \text{inv}_{01} + d_y \cdot \text{inv}_{11})
  \end{aligned}
\end{equation}
For the clamped power term:
\begin{equation}
  \frac{\partial \text{power}}{\partial \text{mahal\_dist}} = \begin{cases} -0.5 & \text{if } \text{power} > -50 \\ 0 & \text{otherwise} \end{cases}
\end{equation}
Tanh activation backward:
\begin{equation}
  \begin{aligned}
    \frac{\partial x_x}{\partial \tilde{x}_x} &= \frac{W}{2} \cdot (1 - \tanh^2(\tilde{x}_x)) \\
    \frac{\partial x_y}{\partial \tilde{x}_y} &= \frac{H}{2} \cdot (1 - \tanh^2(\tilde{x}_y))
  \end{aligned}
\end{equation}

\item \textbf{Gradient for Inverse Covariance Elements}

The gradient w.r.t. inverse covariance (stored as 3 elements):
\begin{equation}
  \begin{aligned}
    \frac{\partial \mathcal{L}}{\partial \text{inv}_{00}} &= \sum_{x,y} \frac{\partial \mathcal{L}}{\partial \mathcal{G}_n} \cdot \mathcal{G}_n \cdot \left(-\frac{1}{2}\right) \cdot d_x^2 \\
    \frac{\partial \mathcal{L}}{\partial \text{inv}_{01}} &= \sum_{x,y} \frac{\partial \mathcal{L}}{\partial \mathcal{G}_n} \cdot \mathcal{G}_n \cdot (-1) \cdot d_x \cdot d_y \\
    \frac{\partial \mathcal{L}}{\partial \text{inv}_{11}} &= \sum_{x,y} \frac{\partial \mathcal{L}}{\partial \mathcal{G}_n} \cdot \mathcal{G}_n \cdot \left(-\frac{1}{2}\right) \cdot d_y^2
  \end{aligned}
\end{equation}

\item \textbf{Gradient for Pre-activation Scales $\tilde{s}_x, \tilde{s}_y$}

Through $\tilde{s} \rightarrow s \rightarrow \Sigma \rightarrow \Sigma^{-1} \rightarrow \text{power}$:
\begin{equation}
  \begin{aligned}
    \frac{\partial \mathcal{L}}{\partial s_x} &= 2s_x \Big(\cos^2\theta \cdot \frac{\partial \mathcal{L}}{\partial \Sigma_{00}} + \cos\theta \sin\theta \\
    &\quad \cdot \frac{\partial \mathcal{L}}{\partial \Sigma_{01}} + \sin^2\theta \cdot \frac{\partial \mathcal{L}}{\partial \Sigma_{11}}\Big)
  \end{aligned}
\end{equation}
\begin{equation}
  \begin{aligned}
    \frac{\partial \mathcal{L}}{\partial s_y} &= 2s_y \Big(\sin^2\theta \cdot \frac{\partial \mathcal{L}}{\partial \Sigma_{00}} - \cos\theta \sin\theta \\
    &\quad \cdot \frac{\partial \mathcal{L}}{\partial \Sigma_{01}} + \cos^2\theta \cdot \frac{\partial \mathcal{L}}{\partial \Sigma_{11}}\Big)
  \end{aligned}
\end{equation}
Exponential backward:
\begin{equation}
  \begin{aligned}
  \frac{\partial \mathcal{L}}{\partial \tilde{s}_x} &= \frac{\partial \mathcal{L}}{\partial s_x} \cdot \exp(\tilde{s}_x) \\
  \frac{\partial \mathcal{L}}{\partial \tilde{s}_y} &= \frac{\partial \mathcal{L}}{\partial s_y} \cdot \exp(\tilde{s}_y)
  \end{aligned}
\end{equation}

\item \textbf{Gradient for Rotation $\theta$}

\begin{equation}
  \begin{aligned}
    \frac{\partial \mathcal{L}}{\partial \theta} &= 2(s_y^2 - s_x^2) \cos\theta \sin\theta \cdot \frac{\partial \mathcal{L}}{\partial \Sigma_{00}} \\
    &\quad + (s_x^2 - s_y^2)(\cos^2\theta - \sin^2\theta) \cdot \frac{\partial \mathcal{L}}{\partial \Sigma_{01}} \\
    &\quad + 2(s_x^2 - s_y^2) \cos\theta \sin\theta \cdot \frac{\partial \mathcal{L}}{\partial \Sigma_{11}}
  \end{aligned}
\end{equation}
\end{enumerate}

\section{Differentiable Light Propagation}
\label{supplementary:light_propagation}

This section provides detailed mathematical formulations and gradient derivations for our \BLASM method.

\subsection{\BLASM Algorithm}

Algorithm~\ref{alg:light_propagation} details the \BLASM method used for hologram reconstruction.

\begin{algorithm}[ht!]
\caption{Bandlimited Angular Spectrum Method}
\label{alg:light_propagation}
\footnotesize
\begin{algorithmic}[1]
\Require $\tilde{U}(f_x, f_y, 0)$: Fourier-domain input field
\Require $\lambda$: wavelength, $d$: propagation distance
\Require $\Delta x$: pixel pitch, $N_x, N_y$: resolution
\Require $a$: aperture size (optional)

\Function{PropagateField}{$\tilde{U}, \lambda, d, \Delta x, N_x, N_y, a$}
    \State $k \leftarrow 2\pi/\lambda$
    \State $L_x \leftarrow N_x \cdot \Delta x$
    \State $L_y \leftarrow N_y \cdot \Delta x$

    \State $\tilde{U}_{\text{out}} \leftarrow$ InitEmpty($N_x, N_y$)

    \ForAll{$(i_x, i_y)$ \textbf{in parallel}}
        \State $f_x \leftarrow (i_x - N_x/2) / L_x$
        \State $f_y \leftarrow (i_y - N_y/2) / L_y$

        \State $f_x^{\max} \leftarrow \frac{1}{\lambda\sqrt{(2d/L_x)^2 + 1}}$
        \State $f_y^{\max} \leftarrow \frac{1}{\lambda\sqrt{(2d/L_y)^2 + 1}}$

        \If{$|f_x| \geq f_x^{\max}$ \textbf{or} $|f_y| \geq f_y^{\max}$}
            \State $\tilde{U}_{\text{out}}[i_x, i_y] \leftarrow 0$
            \State \textbf{continue}
        \EndIf

        \State $k_z^2 \leftarrow k^2 - (2\pi)^2(f_x^2 + f_y^2)$
        \State $k_z \leftarrow \begin{cases} \sqrt{k_z^2} & \text{if } k_z^2 > 0 \\ 0 & \text{otherwise} \end{cases}$

        \State $\cos_\phi, \sin_\phi \leftarrow \cos(k_z d), \sin(k_z d)$
        \State $H \leftarrow \cos_\phi + j \sin_\phi$
        \State $\tilde{U}_{\text{out}}[i_x, i_y] \leftarrow \tilde{U}[i_x, i_y] \cdot H$
    \EndFor

    \If{$a > 0$}
        \ForAll{$(i_x, i_y)$ \textbf{in parallel}}
            \State $dx \leftarrow i_x - N_x/2 + 0.5$
            \State $dy \leftarrow i_y - N_y/2 + 0.5$
            \If{$dx^2 + dy^2 \geq a^2$}
                \State $\tilde{U}_{\text{out}}[i_x, i_y] \leftarrow 0$
            \EndIf
        \EndFor
    \EndIf

    \State \Return $\tilde{U}_{\text{out}}$
\EndFunction
\end{algorithmic}
\end{algorithm}

\subsection{Forward Pass}

\subsubsection{Spatial Frequency Computation}

For a hologram of size $N_x \times N_y$ with pixel pitch $\Delta x$, the spatial frequencies at index $(i_x, i_y)$ are computed as:
\begin{equation}
  \begin{aligned}
    L_x &= N_x \cdot \Delta x, \quad L_y = N_y \cdot \Delta x \\
    f_x(i_x) &= \frac{i_x - N_x/2}{L_x} \\
    f_y(i_y) &= \frac{i_y - N_y/2}{L_y}
  \end{aligned}
\end{equation}
where the zero-frequency component is centered at $(N_x/2, N_y/2)$ following FFT-shift convention.

\subsubsection{Bandlimit Computation}

The maximum spatial frequencies that can propagate without aliasing are computed per-thread:
\begin{equation}
  \begin{aligned}
    f_x^{\max} &= \frac{1}{\lambda\sqrt{(2d/L_x)^2 + 1}} \\
    f_y^{\max} &= \frac{1}{\lambda\sqrt{(2d/L_y)^2 + 1}}
  \end{aligned}
\end{equation}
where $\lambda$ is the wavelength and $d$ is the propagation distance.

\subsubsection{Transfer Function Evaluation}

For spatial frequency $(f_x, f_y)$, the wave vector component along propagation direction is:
\begin{equation}
  k_z^2 = k^2 - (2\pi)^2(f_x^2 + f_y^2)
\end{equation}
where $k = 2\pi/\lambda$ is the wave number. The longitudinal wave vector is:
\begin{equation}
  k_z = \begin{cases}
    \sqrt{k_z^2} & \text{if } k_z^2 > 0 \\
    0 & \text{otherwise}
  \end{cases}
\end{equation}
The transfer function is:
\begin{equation}
  H(f_x, f_y, d) = \begin{cases}
    e^{jk_z d} & \text{if } |f_x| < f_x^{\max}, |f_y| < f_y^{\max} \\
    0 & \text{otherwise}
  \end{cases}
\end{equation}
The complex exponential is evaluated using:
\begin{equation}
  e^{jk_z d} = \cos(k_z d) + j\sin(k_z d)
\end{equation}
computed with hardware-accelerated \texttt{sincosf} or \texttt{sincos} functions.

\subsubsection{Field Propagation}

The propagated field in Fourier domain is:
\begin{equation}
  \tilde{U}(f_x, f_y, d) = \tilde{U}(f_x, f_y, 0) \cdot H(f_x, f_y, d)
\end{equation}
For complex multiplication with input $\tilde{U}_{\text{in}} = \text{real}_{\text{in}} + j \cdot \text{imag}_{\text{in}}$ and transfer function $H = \cos(k_z d) + j\sin(k_z d)$:
\begin{equation}
  \begin{aligned}
    \text{real}_{\text{out}} &= \text{real}_{\text{in}} \cos(k_z d) - \text{imag}_{\text{in}} \sin(k_z d) \\
    \text{imag}_{\text{out}} &= \text{real}_{\text{in}} \sin(k_z d) + \text{imag}_{\text{in}} \cos(k_z d)
  \end{aligned}
\end{equation}

\subsubsection{Aperture Filtering}

When aperture size $a > 0$, circular filtering is applied in a separate kernel pass:
\begin{equation}
  \tilde{U}_{\text{out}}(i_x, i_y, d) = \begin{cases}
    \tilde{U}_{\text{out}}(i_x, i_y, d) & \text{if } (i_x - o_x)^2 + (i_y - o_y)^2 < a^2 \\
    0 & \text{otherwise}
  \end{cases}
\end{equation}
where $(o_x, o_y) = (N_x/2 - 0.5, N_y/2 - 0.5)$ is the centered offset.

\subsection{Backward Pass}

The backward pass computes gradients with respect to the input Fourier field $\tilde{U}(f_x, f_y, 0)$ given gradients of the output $\partial \mathcal{L}/\partial \tilde{U}(f_x, f_y, d)$.

\subsubsection{Complex Conjugate Transfer Function}

The gradient flows through the conjugate transfer function:
\begin{equation}
  \frac{\partial \mathcal{L}}{\partial \tilde{U}(f_x, f_y, 0)} = \frac{\partial \mathcal{L}}{\partial \tilde{U}(f_x, f_y, d)} \cdot H^*(f_x, f_y, d)
\end{equation}
where $H^*(f_x, f_y, d) = e^{-jk_z d}$ is the conjugate, equivalent to backward propagation:
\begin{equation}
\begin{split}
  H^*(f_x, f_y, d) &= \cos(-k_z d) + j\sin(-k_z d) \\
                   &= \cos(k_z d) - j\sin(k_z d)
\end{split}
\end{equation}

\subsubsection{Gradient Complex Multiplication}

For input gradients $\frac{\partial \mathcal{L}}{\partial \tilde{U}_{\text{out}}} = \text{grad}_{\text{real}} + j \cdot \text{grad}_{\text{imag}}$:
\begin{equation}
  \begin{aligned}
    \frac{\partial \mathcal{L}}{\partial \text{real}_{\text{in}}} &= \text{grad}_{\text{real}} \cos(-k_z d) - \text{grad}_{\text{imag}} \sin(-k_z d) \\
    \frac{\partial \mathcal{L}}{\partial \text{imag}_{\text{in}}} &= \text{grad}_{\text{real}} \sin(-k_z d) + \text{grad}_{\text{imag}} \cos(-k_z d)
  \end{aligned}
\end{equation}

\subsubsection{Bandlimiting in Backward Pass}

The same bandlimiting conditions apply:
\begin{equation}
  \frac{\partial \mathcal{L}}{\partial \tilde{U}(f_x, f_y, 0)} = \begin{cases}
    0 & \text{if } |f_x| \geq f_x^{\max} \text{ or } |f_y| \geq f_y^{\max} \\
    \text{computed} & \text{otherwise}
  \end{cases}
\end{equation}
Gradients only flow through physically valid propagating modes within the bandlimit.

\section{Justification for Amplitude Decomposition}
\label{supplementary:decompose}
Although the primitive contribution $\alpha_n \mathbf{c}_n g_n(\mathbf{p})$ is a product, the three factors play distinct roles:
$g_n(\mathbf{p})$ is a spatial kernel determined by the covariance (a function of $\mathbf{p}$, not a free parameter);
$\mathbf{c}_n \in \mathbb{R}^C$ provides the per-channel RGB amplitude;
and $\alpha_n$ is a channel-shared opacity bounded to $[0,1]$ by a sigmoid for optimization stability,
consistent with the standard GaussianImage formulation~\cite{zhang2024gaussianimage}.
Collapsing $\alpha_n$ and $\mathbf{c}_n$ into a single unconstrained amplitude removes this bounded control and reduces the optimizer's effective degrees of freedom.
As shown in \refTbl{ablation_decompose}, the merged variant consistently degrades reconstruction quality, even when matched in parameter count.
\begin{table}[ht!]
    \scriptsize
    \centering
    \setlength{\tabcolsep}{4pt}
    \caption{Ablation on amplitude decomposition (50 DIV2K images). Merging $\alpha$ and $\mathbf{c}$ into a single amplitude $\mathbf{A}$ degrades quality.}
    \label{tbl:ablation_decompose}
    \begin{tabular}{lcccc}
        \toprule
        Variant & Params & PSNR $\uparrow$ & SSIM $\uparrow$ & LPIPS $\downarrow$ \\
        \midrule
        Merged ($\mathbf{A} = \alpha \cdot \mathbf{c}$) & 1.8 M & 29.5 & 0.84 & 0.40 \\
        Merged + matched params & 2.0 M & 29.5 & 0.84 & 0.40 \\
        Ours ($\alpha$, $\mathbf{c}$, $g$) & 2.0 M & \textbf{30.7} & \textbf{0.86} & \textbf{0.33} \\
        \bottomrule
    \end{tabular}
\end{table}

\section{Eyebox Analysis and Single-Viewpoint Reconstruction}
\label{supplementary:eyebox}
\begin{figure}[ht!]
    \centering
    \includegraphics[width=0.99\columnwidth]{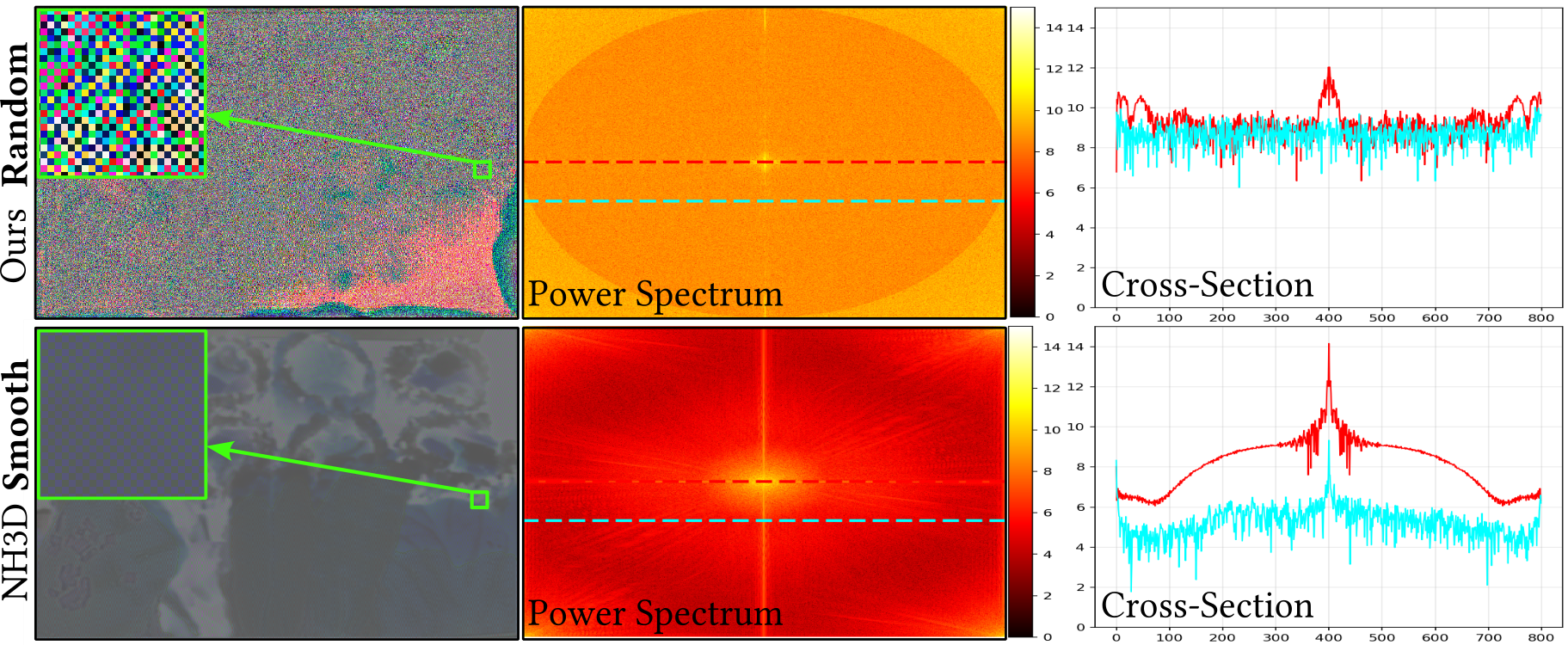}
    \caption{Shifting eyebox result and spectrum analysis.}
    \label{fig:eyebox_analyze}
\end{figure}

\begin{figure}[ht!]
    \centering
    \includegraphics[width=0.99\columnwidth]{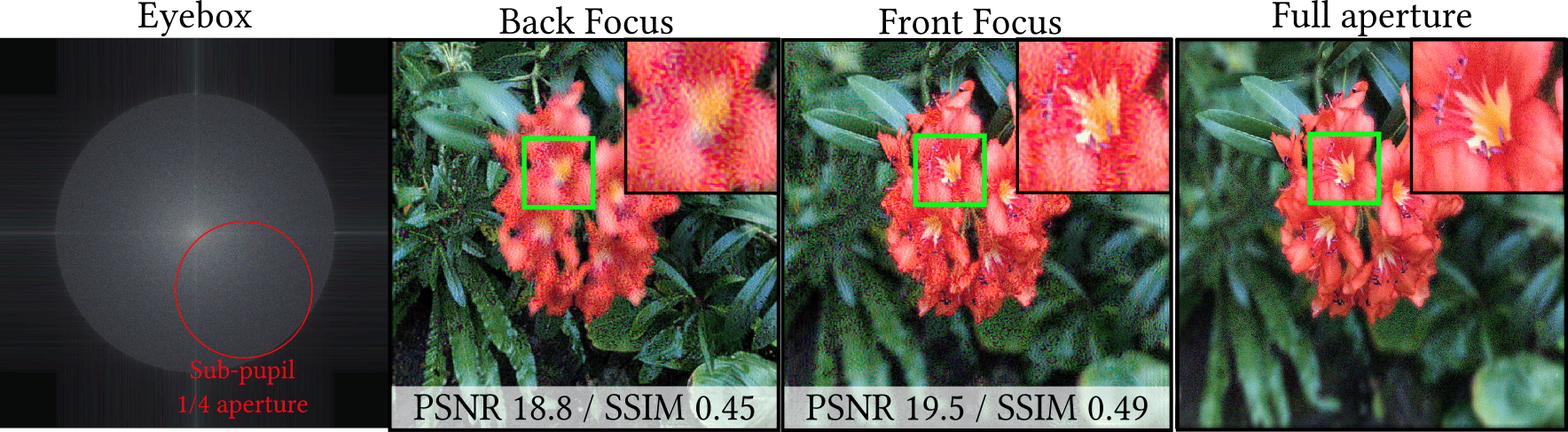}
    \caption{Pupil-shift reconstructions when our representation is optimized under eyebox supervision.}
    \label{fig:eyebox_supervision}
\end{figure}

Our method targets high-fidelity reconstruction at the center of the eyebox.
As shown in \refFig{eyebox_analyze}, our Random \POH exhibits a uniform energy spectrum than Smooth \POH,
similar to typical random \POH~\cite{kavakli2023multicolor}: at pixel-level, scene structures stay recognizable.
In contrast, the smooth \POH from NH3D~\cite{Choi2021Neural3D} shows stronger concentration and sharper peaks.
\setlength{\intextsep}{1.5pt}
\setlength{\columnsep}{5pt}
A uniform spectrum alone, however, does not guarantee a uniform eyebox, it requires \emph{explicit} supervision across pupil positions~\cite{chakravarthula2022pupil, Chu2025RealTime},
so our center-viewpoint objective is complementary to with such methods.
As shown in \refFig{eyebox_supervision}, once our representation is optimized under pupil-shift supervision,
it reconstructs stably across shifted pupil positions, confirming that it extends naturally to the eyebox-expansion setting.

\section{Training Steps Visualization}
\label{supplementary:training_vis_steps}

\begin{figure*}[!thp]
    \centering
    \includegraphics[width=\textwidth]{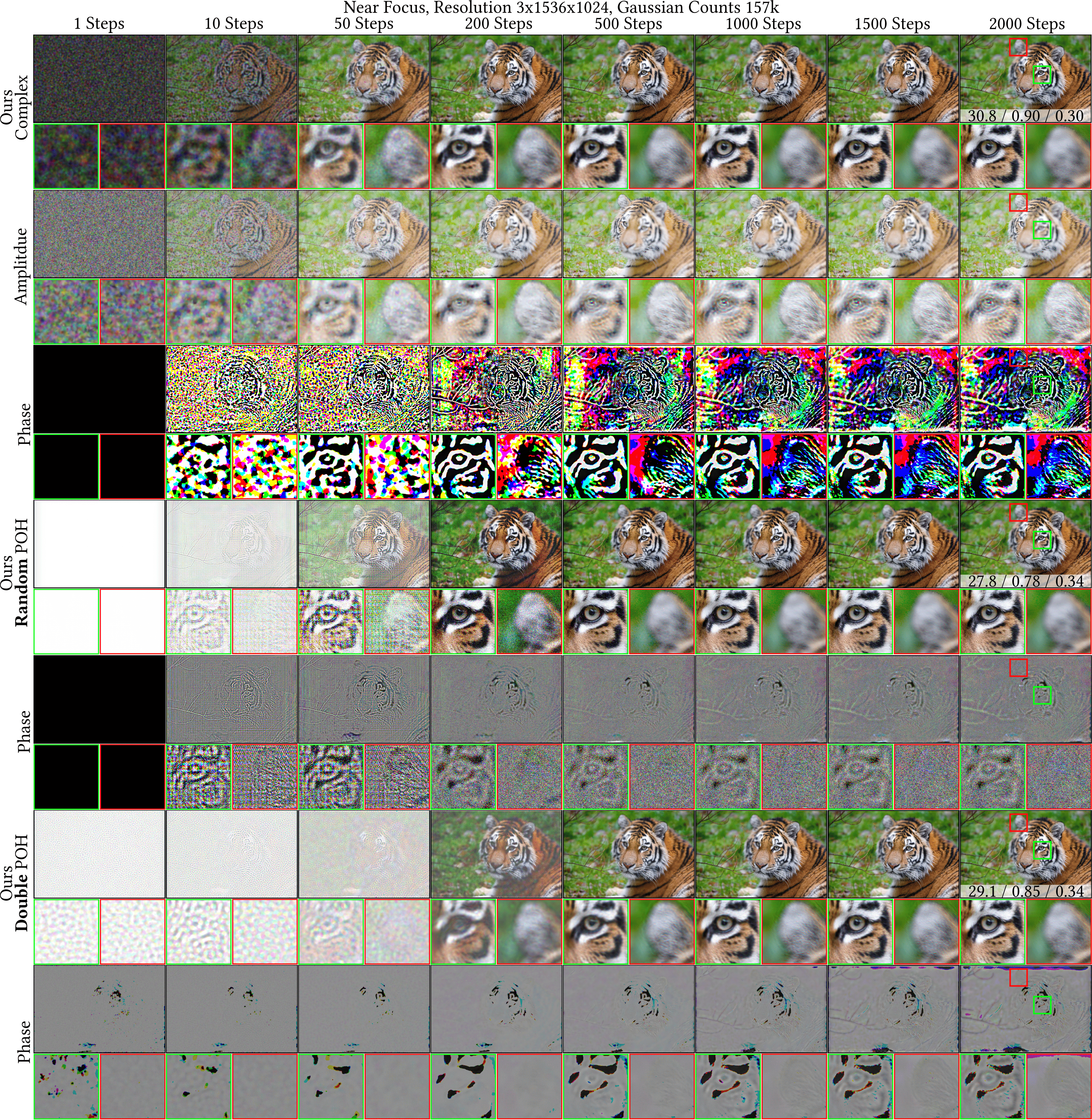}
    \caption{
    Comparison of simulated reconstructions at different training stages; for convenience of space, only the near focal plane is presented.
    The corresponding complex-valued 2D Gaussian hologram and the extracted random and double \POH are shown in parallel.
    Results at 2000 steps are evaluated using PSNR, SSIM, and LPIPS.
    (Source Image: \cite{tiger2015})}
    \label{fig:training_vis_steps}
\end{figure*}

Figure~\ref{fig:training_vis_steps} illustrates the training progression of our method,
showing simulated reconstructions near the focal plane along with the corresponding complex holograms and \POH visualizations.
The image quality becomes stable at around 1000+ steps.

\section{Different Depth Planes Visualization}
\label{supplementary:depth_planes}

\begin{figure*}[!thp]
    \centering
    \includegraphics[width=1\textwidth]{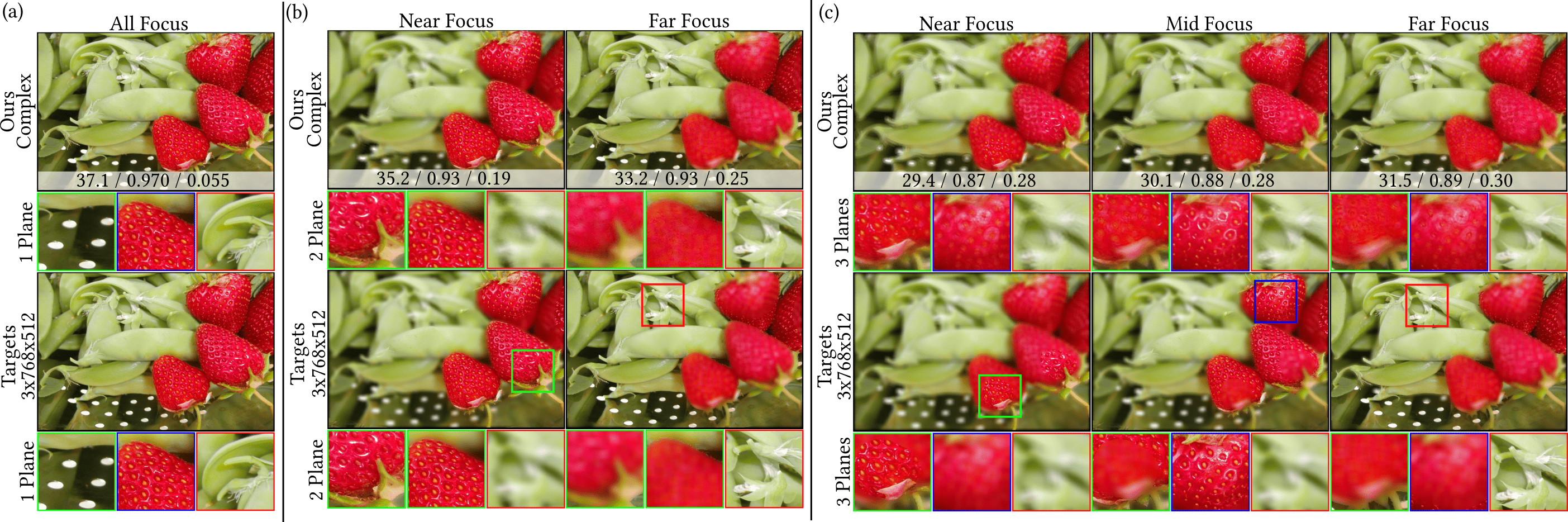}
    \caption{
    Comparison of simulated reconstructions of our method for different depth planes.
    Results are evaluated using PSNR, SSIM, and LPIPS.
    (Source Image: \cite{straw2013})}
    \label{fig:straw_gaussians_different_planes}
\end{figure*}

Figure~\ref{fig:straw_gaussians_different_planes} illustrates the simulated reconstructions of our method across different depth planes ($L = 1, 2, 3$),
showing consistent preservation of fine structures from near to far focus.
The results demonstrate that our representation maintains image fidelity across varying focal depths as reflected by metrics.

\section{Different Propagation Distances Visualization}
\label{supplementary:different_Z}

\begin{figure*}[!thp]
    \centering
    \includegraphics[width=\textwidth]{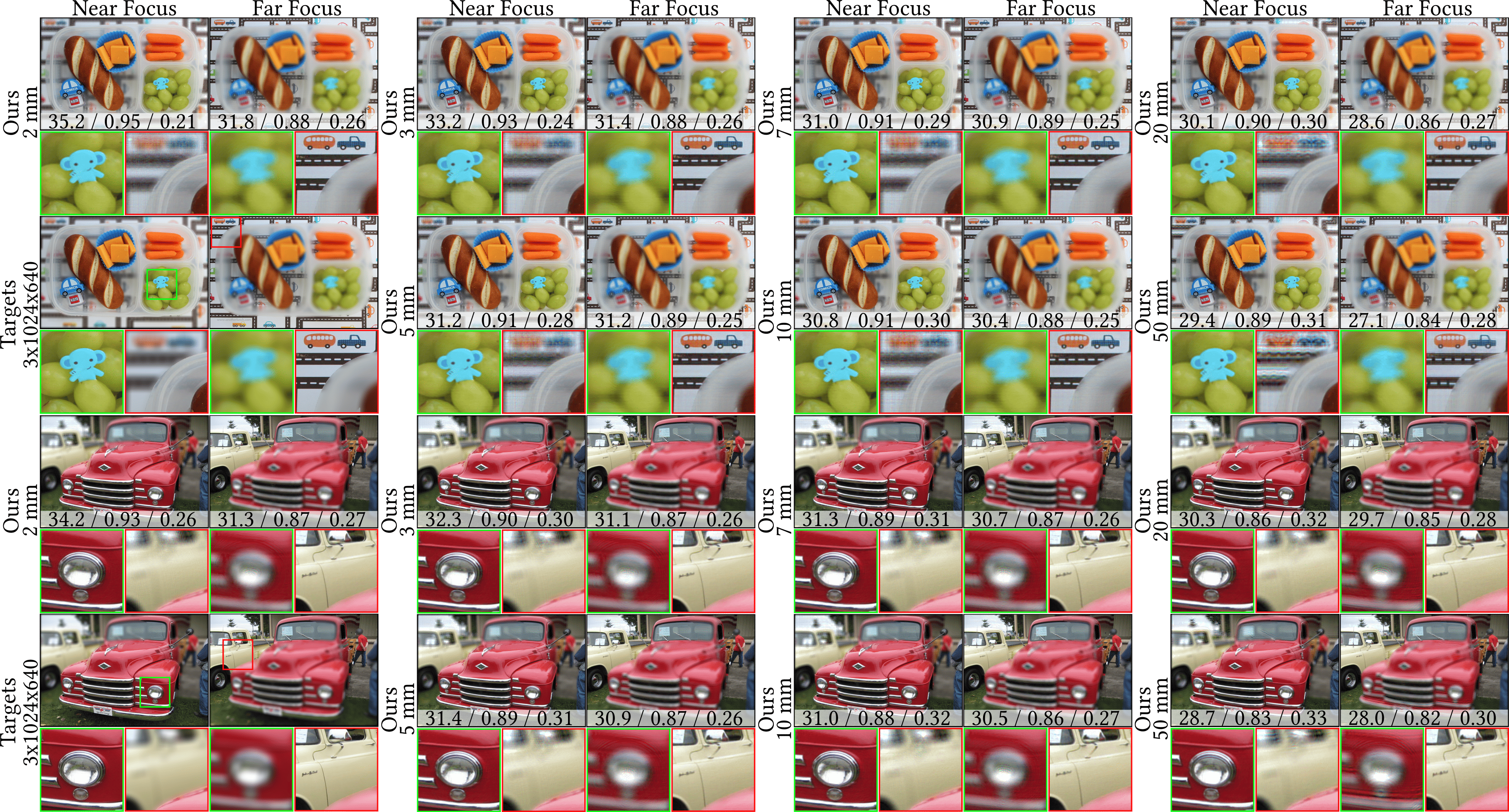}
    \caption{
    Comparison of simulated reconstructions of our method for different propagation distances, ranging from $2 mm$ to $50 mm$ and the volume depth is $4 mm$.
    Results are evaluated using PSNR, SSIM, and LPIPS.
    (Source Image: \cite{AnotherLunch2011, redcar2012})}
    \label{fig:different_Zs}
\end{figure*}

Figure~\ref{fig:different_Zs} illustrates simulated reconstructions of our method at varying propagation distances,
demonstrating consistent preservation of fine structural details from near to far focus.
The results indicate that our representation maintains high image fidelity across a wide range of propagation distances and remains robust even at long distances,
such as $50 mm$, as reflected by the evaluation metrics.

\section{Extra Experimentally Captured Results}
\label{supplementary:extra_exp}
\begin{figure*}[!thp]
    \centering
    \includegraphics[width=1\textwidth]{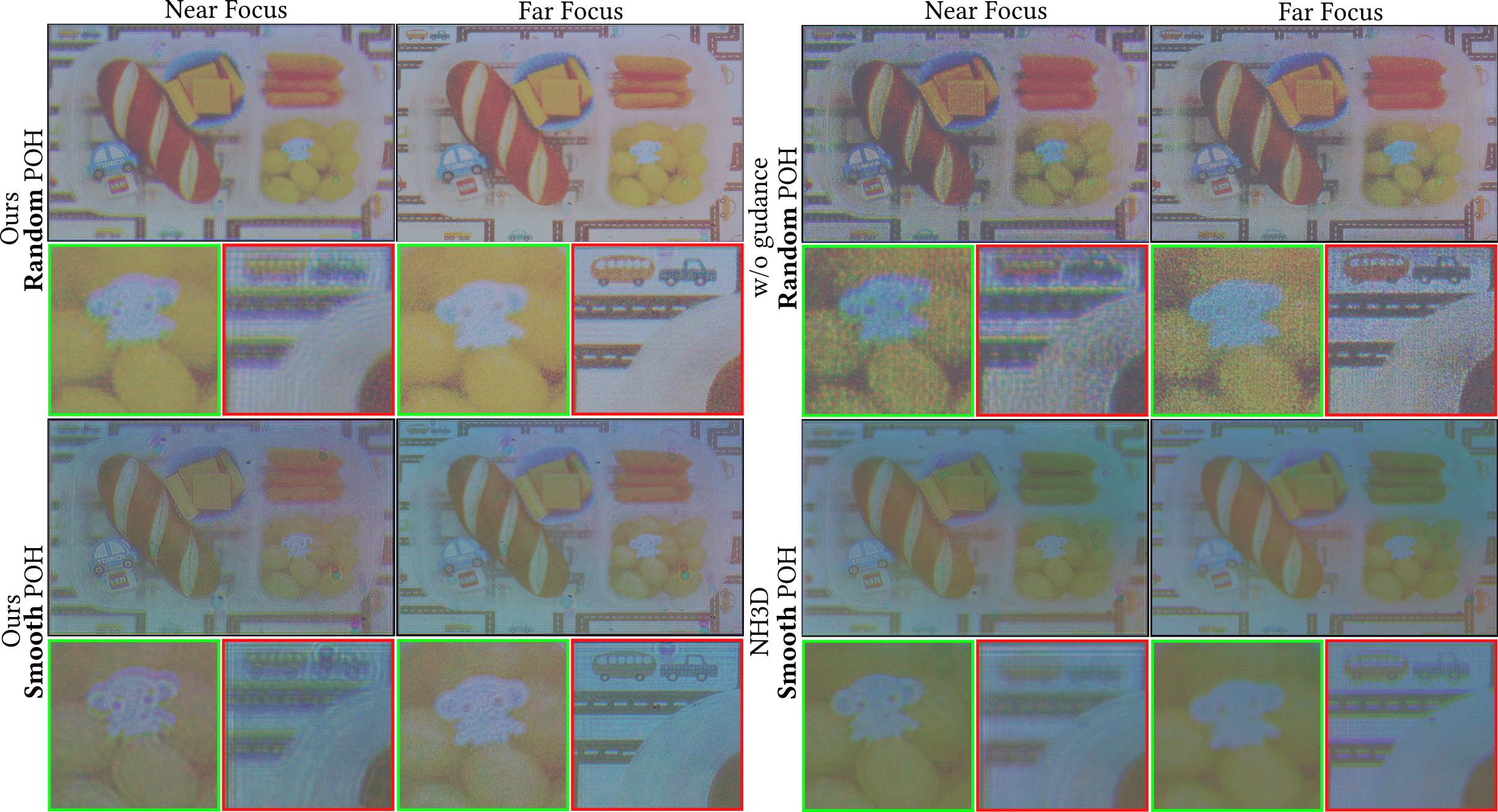}
    \caption{
    Comparison of experimentally captured results of our method with Random  \POH, Smooth \POH, NH3D~\cite{Choi2021Neural3D}, and an independently trained Random  \POH model.
    Results are evaluated using PSNR, SSIM, and LPIPS.
    (Source Image: \cite{AnotherLunch2011})}
    \label{fig:exp_bento}
\end{figure*}

\begin{figure*}[!thp]
    \centering
    \includegraphics[width=1.0\textwidth]{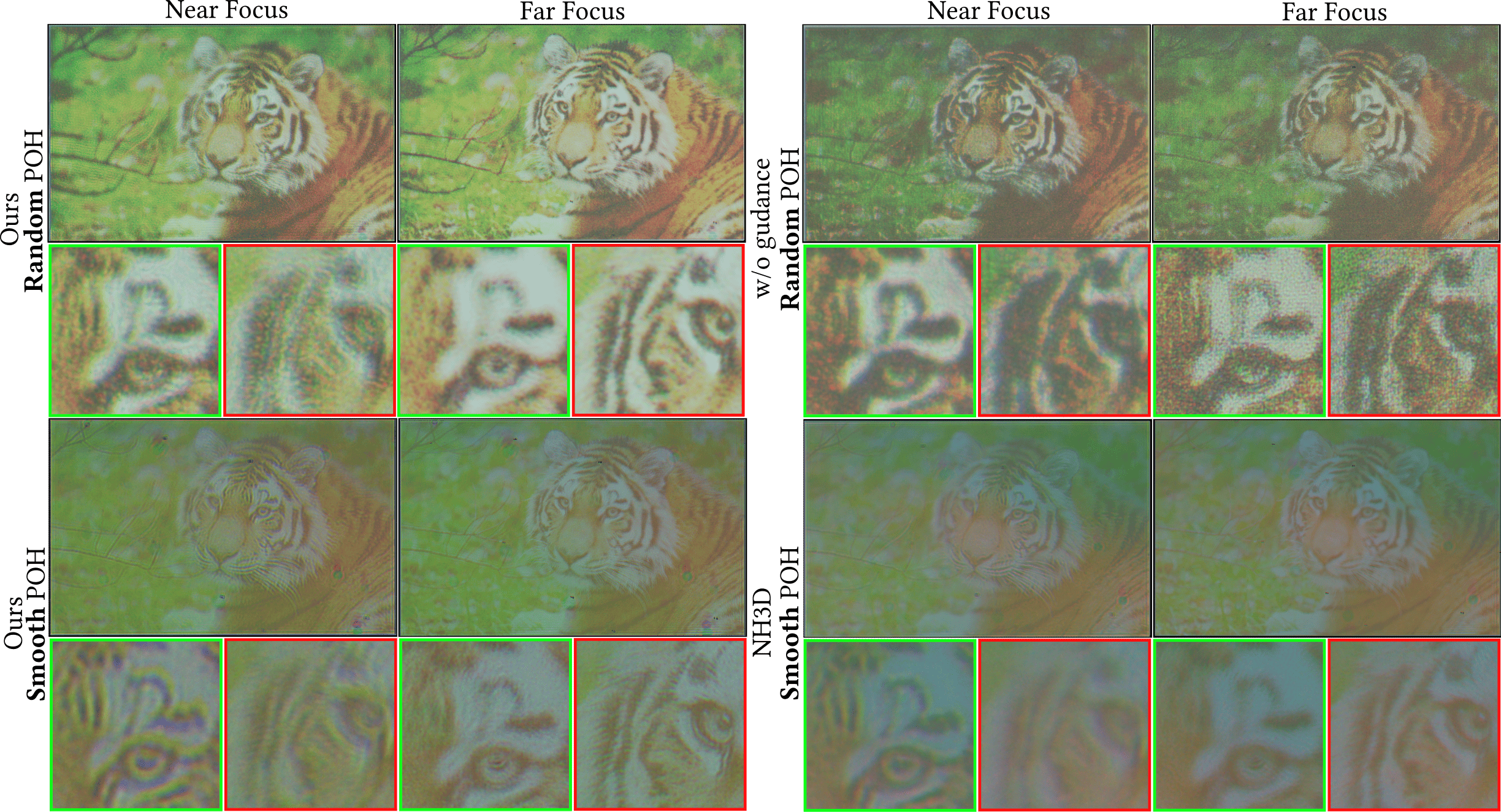}
    \caption{
    Comparison of experimentally captured results of our method with Random  \POH, Smooth \POH, NH3D~\cite{Choi2021Neural3D}, and an independently trained Random  \POH model.
    Results are evaluated using PSNR, SSIM, and LPIPS.
    (Source Image: \cite{tiger2015})}
    \label{fig:exp_tiger}
\end{figure*}

\begin{figure*}[!thp]
    \centering
    \includegraphics[width=1\textwidth]{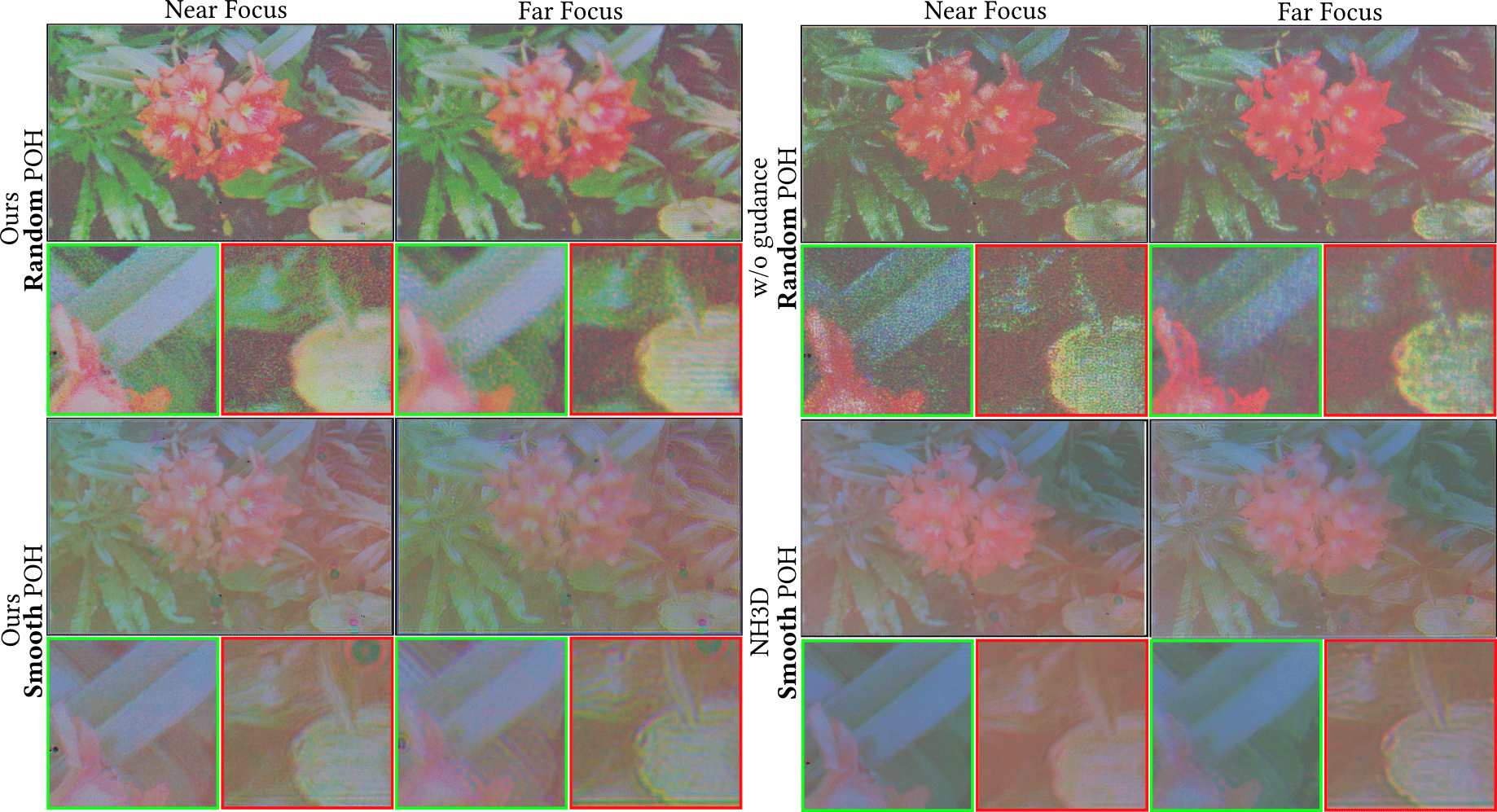}
    \caption{
    Comparison of experimentally captured results of our method with Random  \POH, Smooth \POH, NH3D~\cite{Choi2021Neural3D}, and an independently trained Random  \POH model.
    Results are evaluated using PSNR, SSIM, and LPIPS.
    (Source Image: \cite{mildenhall2019llff})}
    \label{fig:exp_flower}
\end{figure*}

\begin{figure*}[!thp]
    \centering
    \includegraphics[width=1\textwidth]{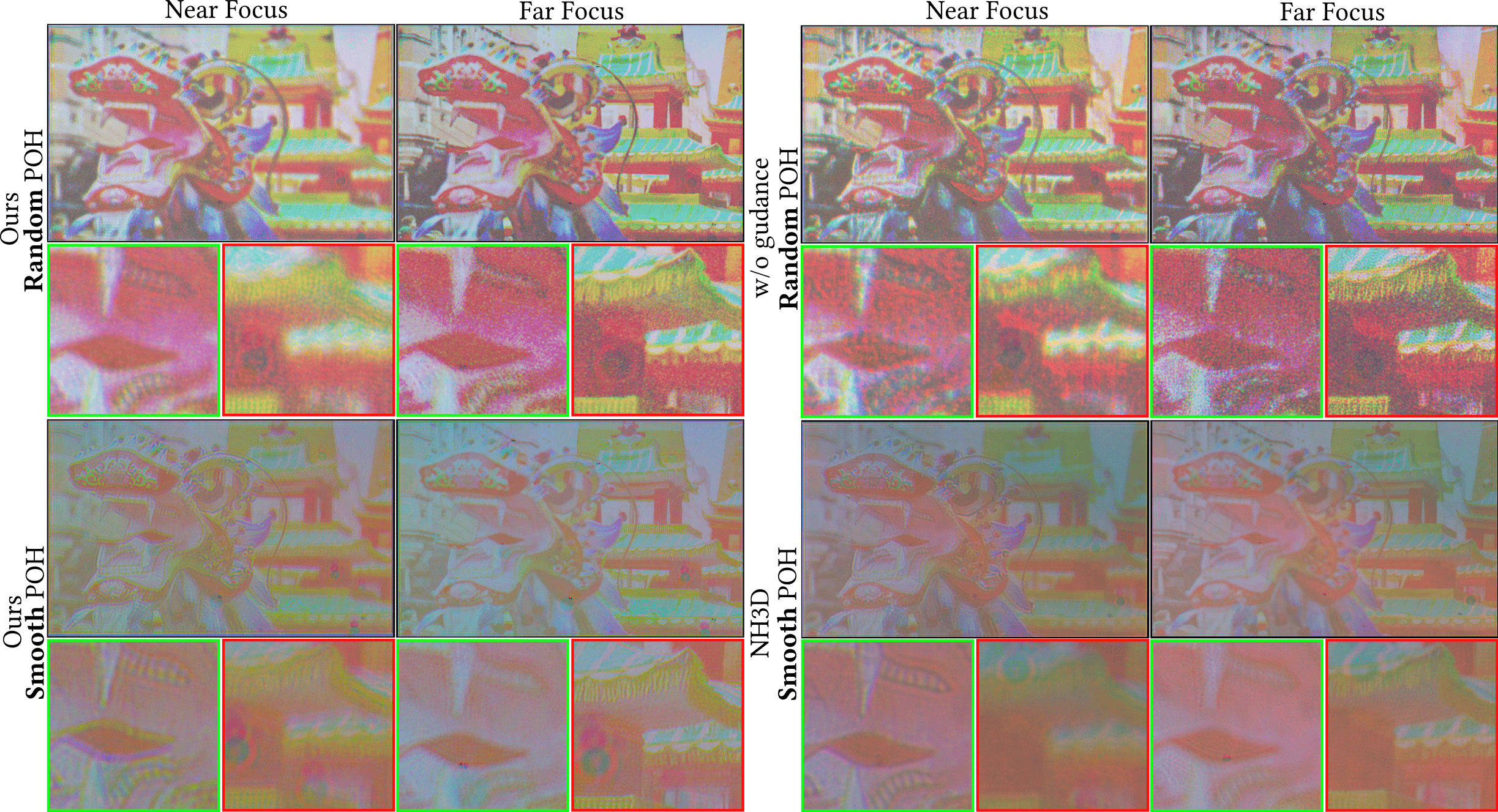}
    \caption{
    Comparison of experimentally captured results of our method with Random  \POH, Smooth \POH, NH3D~\cite{Choi2021Neural3D}, and an independently trained Random  \POH model.
    Results are evaluated using PSNR, SSIM, and LPIPS.
    (Source Image: \cite{dragon2016})}
    \label{fig:exp_dragon}
\end{figure*}

\refFigFull{exp_bento}, \refFigFull{exp_tiger}, \refFigFull{exp_flower}, and \refFigFull{exp_dragon} 
present experimentally captured results across five distinct scenes at resolution of $3 \times 2048 \times 1280$.
Compared to the independently trained Random \POH,
our method achieves an effective suppression of noise without relying on additional time-multiplexing~\cite{choi2022time}, wavelength-multiplexing~\cite{kuo2023multisource, Schiffers2025multiwavelength},
or light-field–based methods~\cite{kim2024holographic, schiffers2023stochastic}, which often costs substantial memory and computational overhead for better image quality.

Although the captures obtained using Smooth \POH also exhibit good image quality with clear focus and defocus,
they suffer from reduced contrast and brightness relative to Random  \POH.
To partially mitigate this degradation, a different set of laser powers was applied during acquisition,
which, however, introduces a noticeable shift in the overall color tone compared with the captures from Random  \POH.

\end{document}